\documentclass[journal,twoside]{IEEEtran}
\usepackage[table]{xcolor}
\usepackage{amsmath,amssymb,amsfonts}
\usepackage{algorithmic}
\usepackage{algorithm}
\usepackage{array}
\usepackage[caption=false,font=normalsize,labelfont=sf,textfont=sf]{subfig}
\usepackage{textcomp}
\usepackage{stfloats}
\usepackage{url}
\usepackage{verbatim}
\usepackage{graphicx}
\usepackage{multirow}
\usepackage{booktabs}
\usepackage{makecell} 
\usepackage{pifont}
\usepackage{cite}
\usepackage{xcolor}
\makeatletter
\let\NAT@parse\undefined
\makeatother
\usepackage[colorlinks=true,
linkcolor=blue,
citecolor=blue,
urlcolor=black,
pdfborder={0 0 1} 
]{hyperref}
\usepackage{orcidlink} 

\def\BibTeX{{\rm B\kern-.05em{\sc i\kern-.025em b}\kern-.08em
    T\kern-.1667em\lower.7ex\hbox{E}\kern-.125emX}}

\begin{document}

\title{Symmetry-Aware 9D Pose Estimation with Sim(3)-Consistent \\ Feature and  Spherical Inception Convolution}

\author{Panfei Cheng\orcidlink{0009-0007-0349-7154}, Hongshan Yu\orcidlink{0000-0003-1973-6766}, Wenrui Chen\orcidlink{0000-0002-6366-7721},\\ Xiaojun Tang, Jian Liu\orcidlink{0000-0003-0604-8024} and Naveed Akhtar\orcidlink{0000-0003-3406-673X},~\IEEEmembership{Member,~IEEE}

\thanks{Panfei Cheng, Hongshan Yu, Wenrui Chen and Jian Liu are with the National
Engineering Research Center for Robot Visual Perception and Control, School of Robotics and Artificial Intelligence, Hunan University, Changsha 410012, China. (email: \href{mailto:chengpf@hnu.edu.cn}{chengpf@hnu.edu.cn}, \href{mailto:yuhongshan@hnu.edu.cn}{yuhongshan@hnu.edu.cn}, \href{mailto: chenwenrui@hnu.edu.cn}{ chenwenrui@hnu.edu.cn}, \href{mailto: jianliu@hnu.edu.cn}{ jianliu@hnu.edu.cn})).}%
\thanks{Xiaojun Tang is with the Beijing Spacecrafts, China Academy of Space Technology, Beijing 100094, China. (email: \href{mailto:xiaojuntang87@castbsc.cn}{xiaojuntang87@castbsc.cn}).}%
\thanks{Naveed Akhtar  is with the School of Computing and Information Systems, The University of Melbourne, Parkville, Victoria 3010, Australia (email: \href{mailto:naveed.akhtar1@unimelb.edu.au}{naveed.akhtar1@unimelb.edu.au}).}%
}


\maketitle
\begin{abstract}
Object pose estimation is a fundamental problem for an agent system to perceive or manipulate objects in images or videos. However, current instance-level methods struggle with generalization to unseen objects. Category-level methods seek to address this, but remain constrained by the complexities of learning in the non-linear Sim(3) space and intra-class variations. 
To address these challenges, We propose an effective method for category-level object pose estimation with two key innovations: (1) A translation/size estimator, featuring a semantic-guided symmetry-aware module 
that leverages  robust generalization capabilities of a large vision model (LVM) to infer symmetry points, resulting in accurate translation and size without shape priors. This result serves as a precomputed cue for rotation estimation, thereby reducing the difficulty of learning in the non-linear Sim(3) space and laying a robust foundation for tackling the inherently more challenging rotation estimation. (2) A feature fusion module, based on our proposed spherical large-kernel inception convolution, fuses semantic features from the LVM with systematically computed geometric features to extract essential pose features from intra-class variations by modeling long-range dependencies without excessive  computational cost. 
Built on these innovations, we achieve SOTA on benchmarks and real-world scenes, while developing a robust robotic picking system capable of handling diverse objects. Our code will be available at the project page: {\hypersetup{urlcolor=blue}\url{https://panfei-cheng.github.io/SSH-Pose}}.

\end{abstract}

\begin{IEEEkeywords}
Object pose estimation, robotic picking, sim(3)-consistent feature, inception convolution.
\end{IEEEkeywords}

\section{Introduction}
\label{sec:introduction}
\IEEEPARstart{O}{bject} pose estimation serves as a critical prerequisite for AR/VR\cite{tang20193d,su2019deep,marchand2015pose}, object manipulation \cite{liu2024domain, yu2023robotic, kim2002object, chen2020edge}, and facilitating human-robot interaction \cite{yu2025category, liang2024adaptive}. Whereas instance-level methods\cite{zhang2025learning,zhou2025canonical,liu2022hff6d} for estimating object poses deliver high accuracy, they need an instance-specific CAD model and inherently lack generalization to unseen objects. To overcome these limitations, research has shifted towards category-level object pose estimation\cite{wang2019normalized,chen2021sgpa,lin2022category,liu2024mh6d,liu2026survey,SinRef6D}. This paradigm  extends 6D object pose estimation\cite{zhou2025multiscale, he2023contourpose} to 9D Pose (3D translation, 3D rotation, and 3D metric size), simultaneously removing the requirement for instance-specific CAD models and enabling generalization to novel instances within known categories. 
Despite these significant advantages in real world and modern industrial manufacturing, the inherent complexity of pose estimation in the non-linear Sim(3) space (Similarity Transformations in 3D space), coupled with diverse intra-class variations, results in far-from-satisfactory performance.

\begin{figure}[t]
    \centering
    \includegraphics[width=\linewidth]{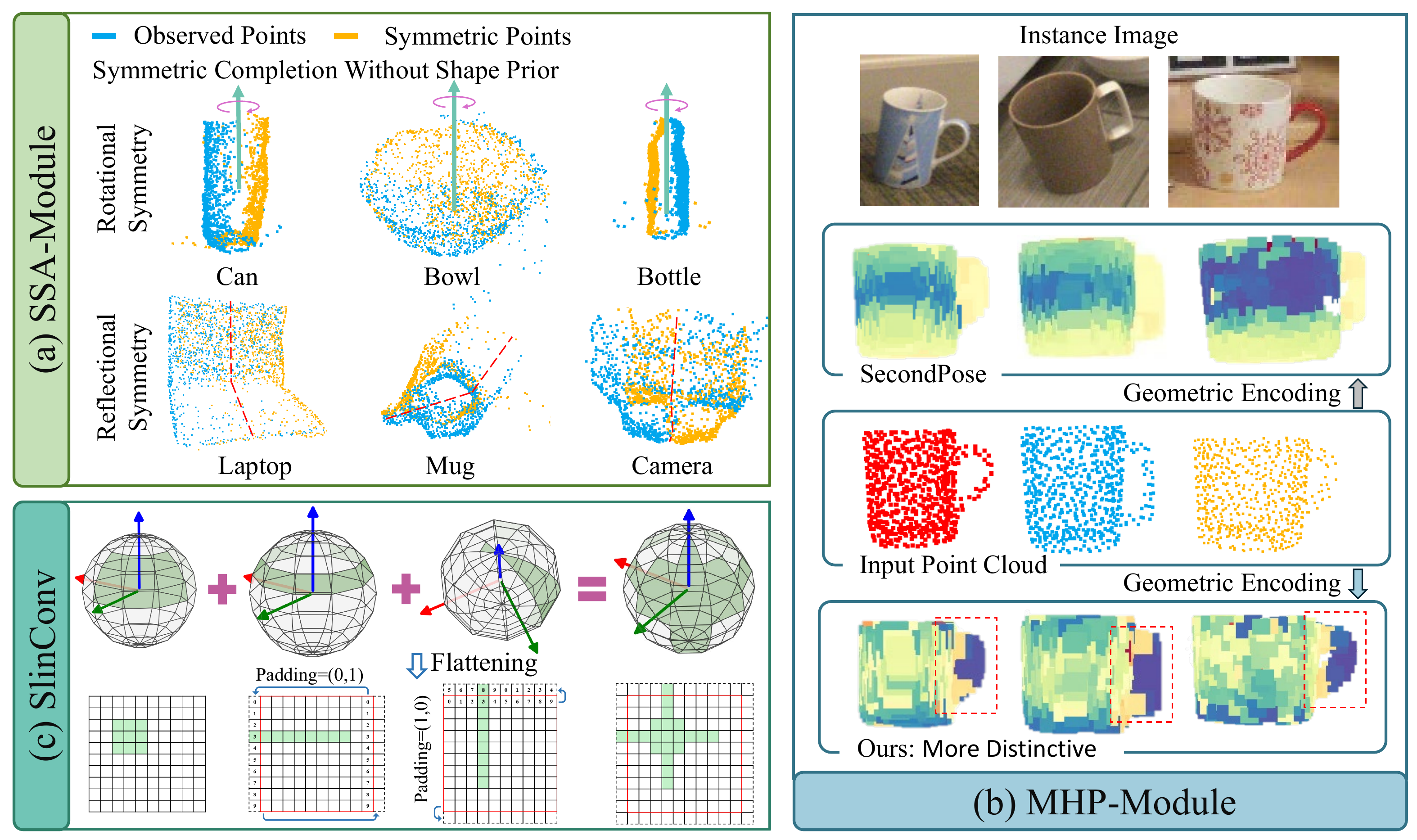}
    \caption{The proposed method generates Rotational-symmetric or Reflectional-symmetric representations of point clouds from a single viewpoint, as shown in (a). For weakly symmetric objects like cameras, we still use symmetry processing, as their translation happens not at the lens axis but at the camera's geometric center, which is closer to its overall symmetric center plane. In (b), our MHP-Module generates more distinctive features as compared to existing methods, e.g., SecondPose\cite{chen2024secondpose}, for the handle of the cup, which contains more orientation information. The concept of Spherical large-kernel inception convolution is illustrated in (c) and detailed in Section \ref{Method}.
}
    \label{fig:introduction}
\end{figure}

A common method to tackle intra-class variation is using shape priors to encapsulate category information\cite{tian2020shape,chen2021sgpa,li2025gce}. However, learning shape priors is time-consuming and struggles to comprehensively encapsulate the structural information of intra-category objects in practice. Therefore, prior-free based methods\cite{liu2023net,lin2024instance,lin2021dualposenet,di2022gpv,zheng2023hs} are becoming increasingly popular. Meanwhile, it has also become particularly important to learn features that are more meaningful for pose estimation without shape priors. The complexities of learning in the nonlinear Sim(3) space and intra-class variations without shape priors make pose estimation more challenging.

Recently, some prior-free methods\cite{ren2025learning,chen2024secondpose,lin2023vi} completely decouple 9D pose estimation into translation, size, viewpoint rotation, and in-plane rotation to tackle the complexity of learning in the nonlinear Sim(3) space. Typically, these methods first estimate the object's translation and size, then translate and scale the object before rotation estimation. However, such a serial pipeline forces the method to prioritize efficiency, making it impossible to adopt more complex networks for extracting more effective features in addressing intra-class variation. Meanwhile, the inaccurate outputs of simple translation/size estimator is used as priors for rotation estimation, which undermines the performance of even elaborately designed decoupled rotation estimation modules. 

This paper follows the decoupled estimation paradigm to tackle nonlinear Sim(3) pose learning complexity, and proposes SSH-Pose, a novel shape prior-free method with simple effective modules for category-level pose estimation.
First, we designed a Semantic-guided \textbf{S}ymmetry-Aware Module (SSA-Module) that combines the strong generalization of DINOv2\cite{oquab2023dinov2} with a simple encoder-decoder (see Fig.~\ref{fig:introduction}(a)). By integrating with PointNet++\cite{qi2017pointnet++} for feature extraction and direct regression, this design enhances translation/size estimation, while providing more accurate priors to simplify pose estimation from Sim(3) to SO(3). Unlike existing completion methods for pose estimation
 \cite{lin2022sar, di2022gpv, zheng2023hs}, our approach neither relies on shape priors nor requires complex network components, except for the reusable DINOv2. The predicted translation/size translate and scale observed points into the object coordinate system.

Next, we project the semantic and geometric features of the points onto the sphere and utilize our proposed \textbf{S}pherical large-kernel inception convolution (Slinconv - see in Fig.~\ref{fig:introduction}(c)) to capture contextual information and extract the high-level spherical feature map without incurring the quadratic computational complexity of Transformers. This design learns more essential features from intra-class variations and maintain lightweight properties. Specifically, the semantic features are obtained from the same DINOv2\cite{oquab2023dinov2} in SSA-Module, while the geometric features are generated by our designed MLP-based \textbf{H}ierarchical Point Pair Feature (MHP) Module. As shown in Fig. \ref{fig:introduction}(b), despite the challenge of intra-class variation, our MHP-Module generates more distinctive features for the cup handle (a key orientation component of the cup) and clearly distinguishes between the root and the middle part of the handle. Besides, the MHP-Module is Sim(3)-consistent and can run in parallel with the translation/size estimation. The spherical feature map generates rotation via a rotation head. Our main contributions are as follows:

\begin{enumerate}
    \item By leveraging the generalization of reusable DINOv2, we designed the SSA-Module for translation and size estimation. This approach eliminates the reliance on shape priors and excessive computation, simplifies pose learning from Sim(3) to SO(3) space, and lays a robust foundation for tackling the inherently more challenging rotation estimation
    \item To track intra-class variations, we propose a novel rotation estimator with two key modules. First, we design a more robust and effective geometric feature generator via the MHP-Module. Second, our Slinconv efficiently models long-range dependencies on the sphere while avoiding the computational overheads of Transformers.
    \item Based on SSH-Pose, we build a robotic picking system. With well-designed translation/size and rotation estimators,  it can robustly and accurately pick category-level unseen objects. Robotic picking experiments are designed to demonstrate the effectiveness of our approach in real-world applications.
   
\end{enumerate}

Besides DINOv2\cite{oquab2023dinov2}, which is commonly used in upstream tasks of object pose estimation, all modules in our network are designed using simple 2D convolution and MLP; yet they achieve state-of-the-art (SOTA) performance on benchmark datasets as well as local real-world challenging scenes. 

\begin{figure*}[!t]
	\centering
	\includegraphics[width=\textwidth]{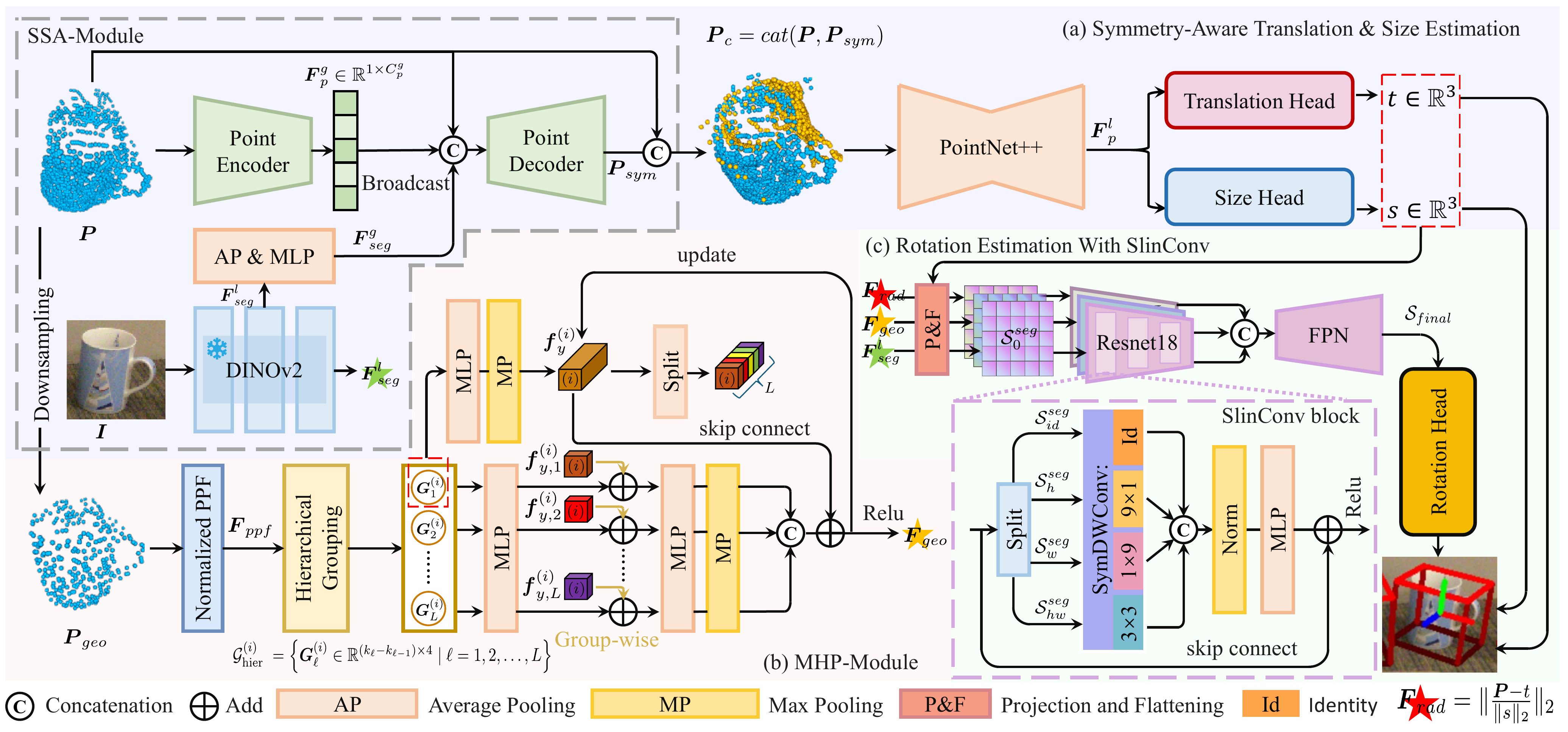}
	\caption{Overview of SSH-Pose, where the three background colors represent its three main components, and the two dashed boxes denote its two sub-modules respectively. We design  (a) Symmetry-aware Translation \& Size Estimation with the SSA-Module to obtain accurate $t$ and $s$. Meanwhile, (b)  MLP-based Hierarchical Point Pair Feature (MHP) Module is proposed to extract effective geometric features. (c) Rotation Estimation with SlinConv fuses semantic features $\boldsymbol{F}_{{seg}}^l$, geometric features $\boldsymbol{F}_{geo}$, and  radius features $\boldsymbol{F}_{rad}$ to estimate rotation. See text for details.}      
	\label{fig:method}

\end{figure*}

\section{RELATED WORK}
We review the related work from three perspectives: shape prior-based methods, shape prior-free methods, and object pose estimation based on 3D reconstruction. Unless otherwise specified, all pose estimation discussed in this section refers to category-level object pose estimation.

\subsection{Shape prior-based methods for pose estimation}
Shape prior-based methods typically use an offline category-level shape prior network, which learns from training-set instance CAD models to generate a category-specific shape prior. As a pioneer, Object-DeformNet\cite{tian2020shape} first deforms the offline-learned shape, uses learned correspondences to get a normalized object coordinate space\cite{wang2019normalized} for observed points, and finally applies the Umeyama algorithm\cite{umeyama2002least} to recover 6D pose and metric size. SAR-Net\cite{lin2022sar} uses shape priors for symmetric completion of observed points, and transforms shape priors to obtain corresponding points for rotation estimation. However, alignment is often non-differentiable and outside deep learning frameworks.
Direct regression methods avoid this by offering an alternative. CenterSnap\cite{irshad2022centersnap} performs object reconstruction and pose estimation using object-centric scene representations. However, its reconstruction requires a complex network and is not directly used for pose estimation. Besides, MH6D\cite{liu2024mh6d} proposes multi-hypothesis consistency learning for category-level pose estimation. Despite good results, the cumbersomeness of category-level shape priors has driven exploration of simpler, more direct solutions.

\subsection{Shape prior-free methods for pose estimation}
Shape prior-free methods are gaining traction in category-level pose estimation owing to their simplicity and usability. These methods can be further categorized into correspondence-based and direct regression approaches.
IST-Net \cite{liu2023net} argues that 3D priors are not the driver of high performance and introduces a deformation network independent of prior knowledge, additionally establishing correspondences to enable pose estimation. For AG-Pose, it leverages an attention module for Instance-Adaptive and Geometric-Aware Keypoint Learning, and similarly estimates pose via correspondence establishment.
On the other hand, DualPoseNet\cite{lin2021dualposenet} adopts dual networks for explicit and implicit direct pose prediction, while enforcing consistency to refine pose estimates. GPV-Pose\cite{di2022gpv} proposes 3DGC to jointly estimate symmetric reconstruction and 9D pose. Building upon GPV-Pose, HS-Pose\cite{zheng2023hs} designs an HS-layer: an efficient module that addresses the insensitivity of GPV-Pose to scale and translation features. Recently, VI-Net\cite{lin2023vi} has been proposed as a novel shape prior-free framework, which improves rotation estimation accuracy by decoupling rotation into viewpoint and in-plane components. Furthermore, SecondPose\cite{chen2024secondpose} enhances VI-Net by integrating an SE(3)-consistent module, which fuses semantic and geometric features for better performance.

\subsection{Pose estimation with 3D object reconstruction}
Owing to occlusions or the inherent limitation of a single viewing angle, the information acquired about an object is invariably partial. This limitation significantly constrains the performance of object pose estimation. An intuitive solution is to complete or reconstruct the object to obtain comprehensive geometric information, which in turn enhances the accuracy of pose estimation. Object-DeformNet\cite{tian2020shape} introduces a deformation network that leverages category-level shape priors to reconstruct the observed object, thereby facilitating object pose estimation. In contrast, IST-Net\cite{liu2023net} contends that shape priors are dispensable during the deformation process. IST-Net\cite{liu2023net} proposes a prior-free 3D object reconstruction network specifically designed for object pose estimation. However, the objects reconstructed by these two types of methods are all in the world coordinate system, losing crucial pose information. GSNet\cite{liu2023gsnet} tackles this issue by utilizing a point cloud completion network based on 3D graph convolution\cite{lin2020convolution} to complete the observed object within the camera coordinate system. However, this completion network suffers from excessive complexity. Recent studies\cite{di2022gpv,zheng2023hs,lin2022sar} have explored approaches based on symmetric inexact completion. While these methods have demonstrated promising results, they still depend on either shape priors or complex network architectures such as 3D graph convolutions.

\noindent \textbf{Discussions:} Overall, whereas SAR-Net\cite{lin2022sar}, GPV-Pose \cite{di2022gpv}, and HS-Pose \cite{zheng2023hs} integrate symmetric reconstruction modules, they require shape priors or rely on intricate geometric features. In contrast, our SSA-Module uses reusable semantic features, eliminating dependence on complex geometric representations while directly boosting pose regression accuracy. The most similar methods are VI-Net~\cite{lin2023vi} and SecondPose~\cite{chen2024secondpose}, which both lack deep geometric features to handle intra-class variations and use 3×3 convolutions, limiting the network’s ability to model long-range pose estimation dependencies. Though Transformers could address this, direct adoption would significantly increase computation and overfitting risks.

\section{Method}
\label{Method}

This paper aims to estimate the 9D pose of diverse instances from known categories, transforming them from the camera coordinate system to the object coordinate system, which is also known as category-level object pose estimation. The inputs are cropped images \(\boldsymbol{I} \in \mathbb{R}^{W \times H \times 3}\) and masked point clouds \(\boldsymbol{P} \in \mathbb{R}^{N \times 3}\) generated by mask-RCNN~\cite{he2017mask}, where $W=210$, $H=210$, $N=2048$. As shown in Fig.~\ref{fig:method}, our SSH-Pose method  consists of three major components.
First, the Symmetry-Aware Translation \& Size Estimation branch comprises a Semantic-guided Symmetry-Aware Module (SSA-Module) that builds upon DINOv2\cite{oquab2023dinov2}. By exploiting symmetric-aware guidance, we estimate accurate translation and size priors, thereby streamlining pose learning from Sim(3) to SO(3) space. This approach establishes a solid foundation for effectively tackling the more challenging rotation estimation.
Then, we propose an MLP-based Hierarchical Point Pair Feature (MHP) Module, which efficiently extracts sim(3)-consistent and robust geometric features to address intra-class variations. 
Finally, rotation estimation with SlinConv is proposed. Leveraging the precomputed translation and size as priors, we reconstruct the masked point clouds $\boldsymbol{P}$ to the origin of the object coordinate system. The geometric, semantic, and radius features are projected onto a sphere, forming a spherical feature map that is highly correlated with rotational information. The SlinConv block is employed to form an encoder-decoder feature pyramid network (FPN) with a ResNet18 structure, which fuses the three spherical features and extracts spherical rotation signals with contextual information. The resulting spherical features are then processed by a rotation head for accurate 3D rotation estimation.

\begin{figure}[t]
    \centering
    \includegraphics[width=\linewidth]{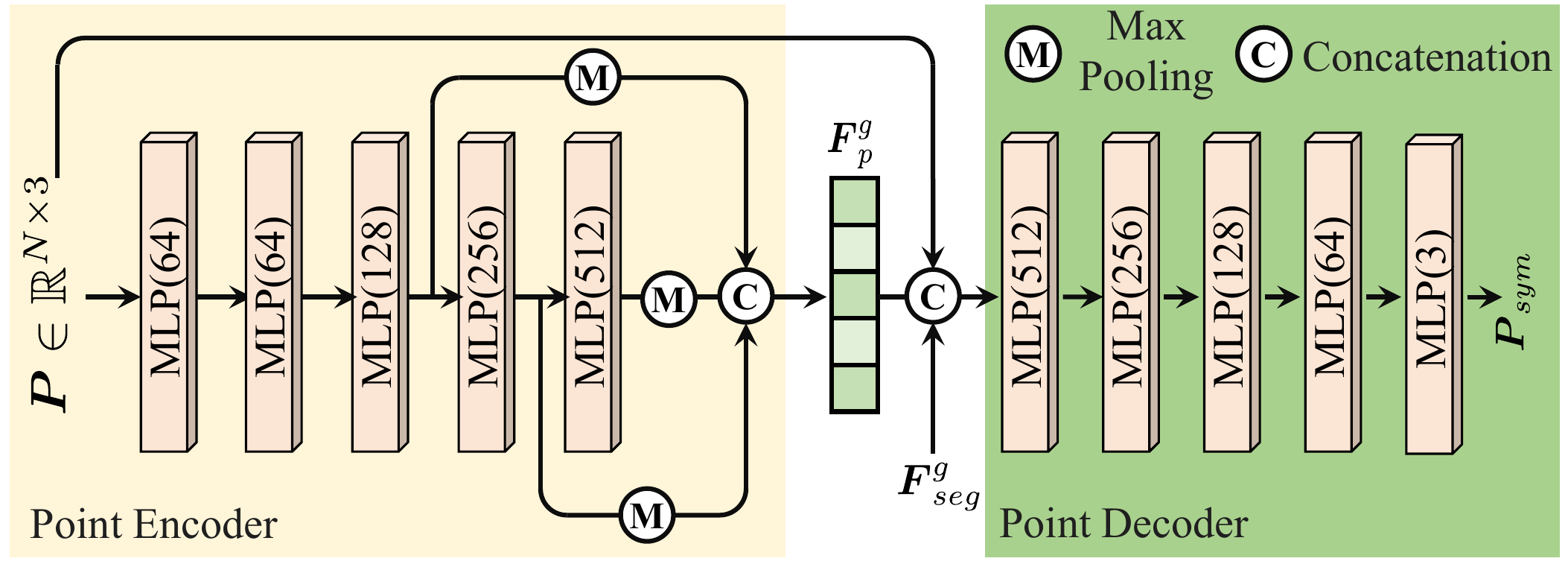}
    \caption{SSA-Module point encoder and decoder architecture.}
    \label{fig:Encoderdecoder}
\end{figure}

\subsection{Symmetry-Aware Translation \& Size Estimation}
Previous works\cite{lin2023vi,chen2024secondpose} also employ separate networks to estimate object translation and size, often treating this as trivial. However, achieving highly accurate translation/size estimation without complex networks remains non-trivial and significantly affects rotation estimation accuracy. This motivates us to revisit this task and develop a more accurate translation/size estimation network. We design a SSA-Module for estimating object translation and metric size. Symmetry awareness is built upon the premise that vast majority of artificial products exhibit global symmetry. On the other hand, sensors typically acquire incomplete data from a single view, while object translation is usually based on the geometric center of the object, and metric size requires measurements in three specific directions, demanding complete point clouds. Unlike SAR-Net~\cite{lin2022sar}, our SSA-Module does not require the pre-preparation of shape priors for each category. Instead, it relies on features from the large semantic model DINOv2\cite{oquab2023dinov2}, which can be reutilized by upstream tasks or other modules. 

As illustrated in Fig.~\ref{fig:method}(a), we generate the global semantic feature \(\boldsymbol{F}_{{seg}}^g \in \mathbb{R}^{1 \times C_{{seg}}^g}\) from the DINOv2 features \(\boldsymbol{F}_{{seg}}^l \in \mathbb{R}^{N_{seg} \times C_{{seg}}^l}\) through average pooling and a simple dimensionality transformation, where $C_{{seg}}^g=1024$, $N_{seg}=225 (15 \times 15)$ and $C_{{seg}}^l=384$. Then, we leverage this global semantic feature \(\boldsymbol{F}_{{seg}}^g\) to guide the symmetric completion of the point cloud \(\boldsymbol{P}\). 
Specifically, we employ a PointNet-like structure\cite{lin2022sar} as the point encoder and point decoder, detailed in Fig. \ref{fig:Encoderdecoder}. The point global features $\boldsymbol{F}_{p}^{g} \in \mathbb{R}^{1 \times C_{p}^{g}}$ generated by the Point Encoder are concatenated with the global semantic features and point-wise coordinates in the feature dimension.  Ultimately, we generate corresponding symmetric points $\boldsymbol{P}_{sym}$ for input points \( \boldsymbol{P}\) and concatenate them to obtain the object's complete points, 
\(\boldsymbol{P}_c \in \mathbb{R}^{2N \times 3}\).

To further process the completed points, we employ a PointNet++\cite{qi2017pointnet++} network with three downsampling steps 
 and generate point features \(\boldsymbol{F}_{\mathrm{p}}^l \in \mathbb{R}^{2N \times 256}\). We then adopt two simple MLP-based output heads\cite{lin2023vi,chen2024secondpose} to estimate the translation \(t \in \mathbb{R}^{3}\) and size \(s \in \mathbb{R}^{3}\), respectively. Notably, even incompletely symmetric objects (e.g., cameras) can benefit from this symmetric structure. Since our estimation of the object's position and size does not rely on extremely fine-grained local details, even inaccurate symmetric completion still facilitates the estimation of position and size.

\subsection{MLP-based Hierarchical Point Pair Feature Module} 

Recently, SecondPose\cite{chen2024secondpose} remodels the underlying object pose estimation process from \{point cloud → canonical space\} to \{point cloud → SE(3)-consistent representation → canonical space\}. However, the SE(3)-consistent representation in the latter relies on a known scale, thus requiring the serialization of size estimation and extraction of geometric feature, which reduces the algorithm's efficiency. Our method remodels this process as \{point cloud → Sim(3)-consistent representation → canonical space\}, thereby enabling the synchronous execution of geometric representation and size estimation. SE(3) comprises rigid transformations of rotation and translation in Euclidean space, while Sim(3) extends this with uniform scaling. This not only improves the algorithm's efficiency but also enhances the robustness of point cloud features.

 MHP-Module is proposed to extract effective geometric features \(\boldsymbol{F}_{geo} \in \mathbb{R}^{N_{g} \times C_{geo}}\) with $N_{g}=300$ and $C_{geo}=96$, as illustrated in Fig. \ref{fig:method}(b). To learn the sim(3)-consistent point features, we first introduce a simple Normalized PPF (Point Pair Feature) block. Specifically, we compute a robust scale factor \(s_{iqr}\) based on the Interquartile range (IQR)\cite{dekking2005modern} method from the input points:
\begin{equation}
d_{IQR} = Q_{0.75}\left( \| \boldsymbol{P}_{geo} \|_2 \right) - Q_{0.25}\left( \| \boldsymbol{P}_{geo} \|_2 \right),
\end{equation}
\begin{equation}
s_{iqr} = Q_{0.75}\left( \| \boldsymbol{P}_{geo} \|_2 \right) + k \cdot d_{IQR},
\end{equation}
where \(\boldsymbol{P}_{geo} \in \mathbb{R}^{N_{g} \times 3}\) is the downsampled points.  $Q_{0.75}$ and $Q_{0.25}$ represent the 75th and 25th percentiles of the data, respectively, and $k=1.5$ is an empirical constant. Then, we compute a sim(3)-consistent PPF feature matrix \(\boldsymbol{F}_{ppf} \in \mathbb{R}^{N_{g} \times N_{g} \times 4}\) for the point cloud \(\boldsymbol{P}_{geo}\), where each element \(f_{i,j} \in \mathbb{R}^{4}\) in \(\boldsymbol{F}_{ppf}\) is defined as below. 

\begin{equation}
f_{i, j}=\left[\tilde{d}_{i, j}, \alpha_{i, j}, \beta_{i, j}, \theta_{i, j}\right],
\label{eq:ppf}
\end{equation}
where, \(\tilde{d}_{i, j}=\left\|{p}_j-{p}_i\right\| / s_{iqr}\) is defined as the normalized Euclidean distance between points \({p}_i \in \boldsymbol{P}_{geo}\) and \({p}_j \in \boldsymbol{P}_{geo}\). The angle \(\alpha_{i, j}\) (or \(\beta_{i, j}\)) are the inferior angle formed by the normal of \({p}_i\) (or \({p}_j\)) and the direction from \({p}_i\) to \({p}_j\), while \(\theta_{i, j}\) signifies the inferior angle between the normals of \({p}_i\) and \({p}_j\). Then, given a strictly increasing list of integers \(\mathcal{K}=\left[k_0, k_1, \ldots, k_L\right]\), we obtain a hierarchical set of neighbour indices \(\mathcal{N}_{\ell} \in \mathbb{Z}^{N_{g} \times\left(k_{\ell}-k_{\ell-1}\right)},\quad \ell=1, \ldots, L\), where \(\mathcal{N}_{\ell}\) contains the indices of the \(k_{\ell-1}+1\) to \(k_{\ell}\)-th nearest points in \(\boldsymbol{P}_{geo}\) for every point in \(\boldsymbol{P}_{geo}\). 

For the sake of convenience and clarity in our presentation, subsequent steps of feature processing will be explicitly exemplified using the i-th point in the point cloud; other points will undergo identical operations.
We obtain the hierarchical normalized PPF feature groups (patch) \(\mathcal{G}_{\text {hier }}^{(i)}=\left\{\boldsymbol{G}_{\ell}^{(i)} \in \mathbb{R}^{\left(k_{\ell}-k_{\ell-1}\right) \times 4} \mid \ell=1,2, \ldots, L\right\}\) by indexing from \(\boldsymbol{F}_{ppf}\).
Then, we employ an MLP to perform dimension transformation on the nearest PPF feature group $\boldsymbol{G}_{1}^{(i)}$, and apply max-pooling within the group to obtain intermediate geometric features $\boldsymbol{f}_{y}^{(i)}\in \mathbb{R}^{1\times C_{y}^{geo}}$, where $y$ denotes the $y$-th time this feature is obtained. Subsequently, $\boldsymbol{f}_{y}^{(i)}$ is split into $L$ parts along the feature dimension. Each part, denoted as $\boldsymbol{f}_{y,\ell}^{(i)}$, is added to the PPF feature group processed by the MLP. This addition does not follow the broadcasting mechanism; instead, group-wise addition is performed. Specifically, each feature part $\boldsymbol{f}_{y,\ell}^{(i)}$ is grouped in the same manner as $\boldsymbol{G}_{\ell}^{(i)}$ before the addition operation. Here, the PPF feature group acts as positional encoding. Then, these split features are concatenated after being dimensionally expanded by an MLP and pooled within groups. This concatenated result is used to update the intermediate features after skip connection, where the update operation is performed only once by default. After the features of all points in \(\boldsymbol{P}_{geo}\) are updated, the above operations are repeated, ultimately obtaining the geometric features \(\boldsymbol{F}_{geo}\) through a ReLU function.

Through the MHP-Module, we achieve parallel processing of translation/size estimation and extraction of geometric features, enhancing their ability to handle intra-class variations.

\begin{algorithm}[!t]
    \caption{SymDWConv}
    \label{SymDWConv}
    \renewcommand{\algorithmicrequire}{\textbf{Input:}}
    \renewcommand{\algorithmicensure}{\textbf{Output:}}
    
    \begin{algorithmic}[1]
        \REQUIRE $x\in \mathbb{R}^{h \times w \times c}$, $k_1, k_2\in \mathbb{N}$ \textcolor{blue}{\COMMENT{$k_1$,$k_{2}$ is the kernel size}}
        \ENSURE $y\in \mathbb{R}^{h \times w \times c}$    
        
        \STATE  $x_{padding} \gets \text{PaddingFunction}(x,m=k_1//2, n=k_2//2)$
        \STATE  $x_{1} \gets \text{DWConv}^{c \rightarrow c}_{k_1 \times k_2}(x_{padding})$ \textcolor{blue}{\COMMENT{depthwise convolution}}
        \STATE  $x_{2} \gets \text{DWConv}^{c \rightarrow c}_{k_1 \times k_2}(\text{FLIP}(x_{padding}, \text{dim}=1))$
        \STATE  $y \gets \text{Max}(x_{1}, \text{FLIP}(x_{2}, dim=1))$ \textcolor{blue}{\COMMENT{element-wise max}}
        
        \RETURN $y$
    \end{algorithmic}

\end{algorithm}

\subsection{Rotation Estimation With SlinConv}



First, We project the semantic features $\boldsymbol{F}_{{seg}}^l$, geometric features $\boldsymbol{F}_{{geo}}$, and radius features $\boldsymbol{F}_{rad} = \| \frac{\boldsymbol{P} - t}{\|s\|_2} \|_2$ onto the sphere based on their coordinates transformed by $t$ and $s$. The spherical surface is divided into \(H_0 \times W_0\) bins (\(H_0=W_0=64\)). For features in the same bin, we select those farther from the origin; empty bins are initialized to zero. This process yields three flattened spherical surface feature representations: $\mathcal{S}_{0}^{seg} \in \mathbb{R}^{H_0 \times W_0 \times C_{seg}^l}$, $\mathcal{S}_{0}^{geo} \in \mathbb{R}^{H_0 \times W_0 \times C_{geo}}$ and $\mathcal{S}_{0}^{rad} \in \mathbb{R}^{H_0 \times W_0 \times 1}$.

After obtaining spherical features, works like \cite{chen2024secondpose,lin2023vi} employed a feature pyramid built from 3x3 convolutions. However, 3x3 convolutions lack long-range dependency modeling capability\cite{yu2024inceptionnext}, so their performance is surpassed by Transformers, which in turn introduce quadratic computational complexity issues. To enhance the capability of perceiving rotation information from the sphere under intra-class variation challenges,  we propose a SlinConv block that incorporates context modeling capabilities while minimizing computational overhead.

The overall structure of our  rotation estimation Encoder-Decoder is the FPN with ResNet18. The key difference from existing networks lies in our design of a SlinConv block to replace the 3×3 convolution. Specifically, for the input feature map, e.g., $\mathcal{S}_{0}^{seg}$, we split it along the feature dimension into four distinct partitions: $\mathcal{S}_{hw}^{seg}$, $\mathcal{S}_{h}^{seg}$, $\mathcal{S}_{w}^{seg}$, $\mathcal{S}_{id}^{seg}$. Among them, the feature dimension of $\mathcal{S}_{hw}^{seg}$, $\mathcal{S}_{h}^{seg}$ and $\mathcal{S}_{w}^{seg}$ is set to $C_{seg}^{hw}=C_{seg}^{h}=C_{seg}^{w}=r_gC_{seg}$, where $r_g=0.125$. The feature dimension of the remaining partition is given by $C_{seg}^{id}=(1-3r_g)C_{seg}$. Subsequently, the split features are processed using the parallel SymDWConv module, as illustrated in Algorithm \ref{SymDWConv} and \ref{padding}, where Algorithm \ref{padding} serves to align the differences between spherical convolution and 2D convolution. For \( \mathcal{S}_{hw}^{seg} \), we use a small square kernel size with a default of \( 3 \times 3 \). In contrast, for \( \mathcal{S}_{h}^{seg} \) and \( \mathcal{S}_{w}^{seg} \), we utilize a band kernel size, with default settings of \( 9 \times 1 \) and \( 1 \times 9 \), respectively. Through these operations, we can perceive a larger context range without significantly increasing the computational load. Ultimately, as shown in Fig. \ref{fig:method}(c), we concatenate the features processed by SymDWConv and perform  simple transformations.
Parallel ResNet18 modules with SlinConv blocks are employed to simultaneously process \( \mathcal{S}_{0}^{seg} \), \( \mathcal{S}_{0}^{geo} \), and \( \mathcal{S}_{0}^{rad} \). The output results are concatenated, and a similar inverted FPN is used as a decoder to generate the final feature map $\mathcal{S}_{final}$.  The structures of the encoding and decoding pyramids 
consist of three layers, with each layer connected through standard downsampling or upsampling layers. Finally, we employ the rotation head\cite{lin2023vi} to estimate the rotation matrices. The rotation head obtains the final rotation through voting after performing row-wise and column-wise max pooling on the finally obtained 2D feature map, which echoes our band convolution kernel.

\begin{algorithm}[!t]
    \caption{PaddingFunction}
    \label{padding}
    \renewcommand{\algorithmicrequire}{\textbf{Input:}}
    \renewcommand{\algorithmicensure}{\textbf{Output:}}
    
    \begin{algorithmic}[1]
        \REQUIRE $x\in \mathbb{R}^{h \times w \times c}$, $m,n\in \mathbb{N}$ 
        \ENSURE $x^{padding}\in \mathbb{R}^{(h+2m) \times (w+2n) \times c}$    
        
        \STATE $x_{11} \gets x[0:m, 0:w//2,:]$
        \STATE $x_{13} \gets x[h-m:h,0:w//2,:]$
        \STATE $x_{12} \gets x[0:m,w//2:w,:]$
        \STATE $x_{14} \gets x[h-m:h,w//2:w,:]$

        \IF {$m > 1$}
            \STATE $x_{11}, x_{12} \gets \text{FLIP}(x_{11},0),\ \text{FLIP}(x_{12},0)$
            \STATE $x_{13}, x_{14} \gets \text{FLIP}(x_{13},0),\ \text{FLIP}(x_{14},0)$
        \ENDIF 
        
        \STATE $x_{top} \gets \text{CONCATENATE}(x_{12}, x_{11}, \text{dim}=1)$
        \STATE $x_{bottom} \gets \text{CONCATENATE}(x_{14}, x_{13}, \text{dim}=1)$
        \STATE $x_{b}^{t} \gets \text{CONCATENATE}(x_{top}, x, x_{bottom}, \text{dim}=0)$ 

        \STATE $x_{left},x_{right} \gets x[:, 0:n, :]$, $x[:, w - n:w, :]$
        \STATE $x^{padding} \gets \text{CONCATENATE}(x_{right}, x_{b}^{t}, x_{left}, \text{dim}=1)$
        
        \RETURN $x^{padding}$
    \end{algorithmic}
\end{algorithm}

\subsection{Loss Function}
We train two separate networks for translation-size and rotation. This is in-line with previous related methods \cite{chen2024secondpose,lin2023vi}. The translation-size network utilizes L1 loss for translation, size, and symmetric point cloud as follows.
\begin{equation}
L_{ts} = |t_{pred} - t_{gt}| + |s_{pred} - s_{gt}| + |\boldsymbol{P}_{sym}^{pred} - \boldsymbol{P}_{sym}^{gt}|,
\end{equation}
For rotation estimation, we predict the full 9D rotation matrix and apply L1 loss.
\begin{equation}
L_R = |R_{pred} - R_{gt}|.
\end{equation}

Points are centered/normalized using $t_{gt}$ and $s_{gt}$ during training, but predictions enable this normalization at inference.

{\setlength{\tabcolsep}{6pt}
\begin{table*}[t]
\caption{Quantitative Evaluation: Our Method vs. SOTA on NOCS-REAL275 and NOCS-CAMERA25. Corr denotes correspondence-based methods. DR denotes direct regression methods. `*' denotes the CATRE\cite{liu2022catre} IoU metrics.} 
\label{tab:1}
\centering
\begin{tabular}{@{\hskip\tabcolsep}c|c|c|*{5}{c}|*{5}{c}@{\hskip\tabcolsep}}
\toprule
\multicolumn{2}{c|}{\multirow{2}{*}[-0.5\normalbaselineskip]{\makecell{Method}}} & 
\multirow{2}{*}[-0.5\normalbaselineskip]{\makecell{Shape\\Priors}} & 
\multicolumn{5}{c|}{NOCS-REAL275 (mAP (\%))} & 
\multicolumn{5}{c}{NOCS-CAMERA25 (mAP (\%))} \\ 
\cmidrule(lr){4-8} \cmidrule(lr){9-13}
\multicolumn{2}{c|}{} & & $IoU_{75^*}$ & $5^\circ2\text{cm}$ & $5^\circ5\text{cm}$ & $10^\circ2\text{cm}$ & $10^\circ5\text{cm}$ & $IoU_{75^*}$ & $5^\circ2\text{cm}$ & $5^\circ5\text{cm}$ & $10^\circ2\text{cm}$ & $10^\circ5\text{cm}$ \\ 
\midrule
\multirow{6}{*}{Corr} 
& SPD\cite{tian2020shape} & \checkmark & 27.0 & 19.3 & 21.4 & 43.2 & 54.1 & 46.9 & 54.3 & 59.0 & 73.3 & 81.5 \\
& SAR-Net\cite{lin2022sar}     & \checkmark & --   & 31.6 & 42.3 & 50.3 & 68.3 & --   & 66.7 & 70.9 & 75.3 & 80.3 \\
& SGPA\cite{chen2021sgpa}        & \checkmark & 37.1 & 35.9 & 39.6 & 61.3 & 70.7 & 69.1 & 70.7 & 74.5 & \underline{82.7} & 88.4 \\
& DPDN\cite{lin2022category}        & \checkmark & --   & 46.0 & 50.7 & 70.4 & 78.4 & --   & --   & --   & --   & --   \\
& GPT-COPE\cite{zou2023gpt}    & \checkmark & -- & 45.9 & 53.8 & 63.1 & 77.7 & -- & 70.4 & 76.5 & 81.3 & 88.7 \\
& GCE-Pose\cite{li2025gce}    & \checkmark & --   & \underline{57.0} & \underline{65.1} & \underline{75.6} & \underline{86.3} & --   & --   & --   & --   & --   \\ 
\cmidrule(lr){2-13} 
\multirow{2}{*}{DR} 
& CenterSnap-R\cite{irshad2022centersnap}  & \checkmark & -- & -- & 29.1 & -- & 64.3 & --   & --   & 66.2   & --   & 81.3   \\
& MH6D\cite{liu2024mh6d} & \checkmark & -- & 53.0 & 61.1 & 72.0 & 82.0 & --   & --   & --   & --   & --   \\
\midrule
\midrule
\multirow{2}{*}{Corr} 
& IST-Net\cite{liu2023net}     & \ding{55}  & --   & 47.5 & 53.4 & 72.1 & 80.5 & --   & 47.5 & 53.4 & 72.1 & 80.5 \\
& AG-Pose\cite{lin2024instance}     & \ding{55}  & --   & 54.7 & 61.7 & 74.7 & 83.1 & --   & \textbf{77.8} & \underline{82.8} & \textbf{85.5} & \textbf{91.6} \\
\cmidrule(lr){2-13} 
\multirow{6}{*}{DR} 
& DualPoseNet\cite{lin2021dualposenet} & \ding{55}  & 30.8 & 29.3 & 35.9 & 50.0 & 66.8 & 71.7 & 64.7 & 70.7 & 77.2 & 84.7 \\
& GPV-Pose\cite{di2022gpv}    & \ding{55}  & --   & 32.0 & 42.9 & --   & 73.3 & --   & 72.1 & 79.1 & --   & 89.0 \\
& HS-Pose\cite{zheng2023hs}     & \ding{55}  & --   & 46.5 & 55.2 & 68.6 & 82.7 & --   & 73.3 & 80.5 & 80.4 & 89.4 \\
& VI-Net\cite{lin2023vi}      & \ding{55}  & 48.3 & 50.0 & 57.6 & 70.8 & 82.1 & \underline{79.1} & 74.1 & 81.4 & 79.3 & 87.3 \\
& CLIPose\cite{lin2024clipose} & \ding{55}  & -- & 48.7& 58.3 & 70.4 & 85.2 & -- & 74.8 & 82.2 & 82.0 & 90.6 \\
& SecondPose\cite{chen2024secondpose}  & \ding{55}  & \underline{49.7} & 56.2 & 63.6 & 74.7 & 86.0 & -- & -- & -- & -- & -- \\
& Ours        & \ding{55}  & \textbf{51.7} & \textbf{59.3} & \textbf{66.3} & \textbf{77.2} & \textbf{87.4} & \textbf{79.6} & \underline{76.5} & \textbf{83.5} & \underline{82.7} & \underline{91.4} \\ 
\bottomrule
\end{tabular}
\end{table*}}

\begin{figure*}[!t]
	\centering
	\includegraphics[width=\textwidth]{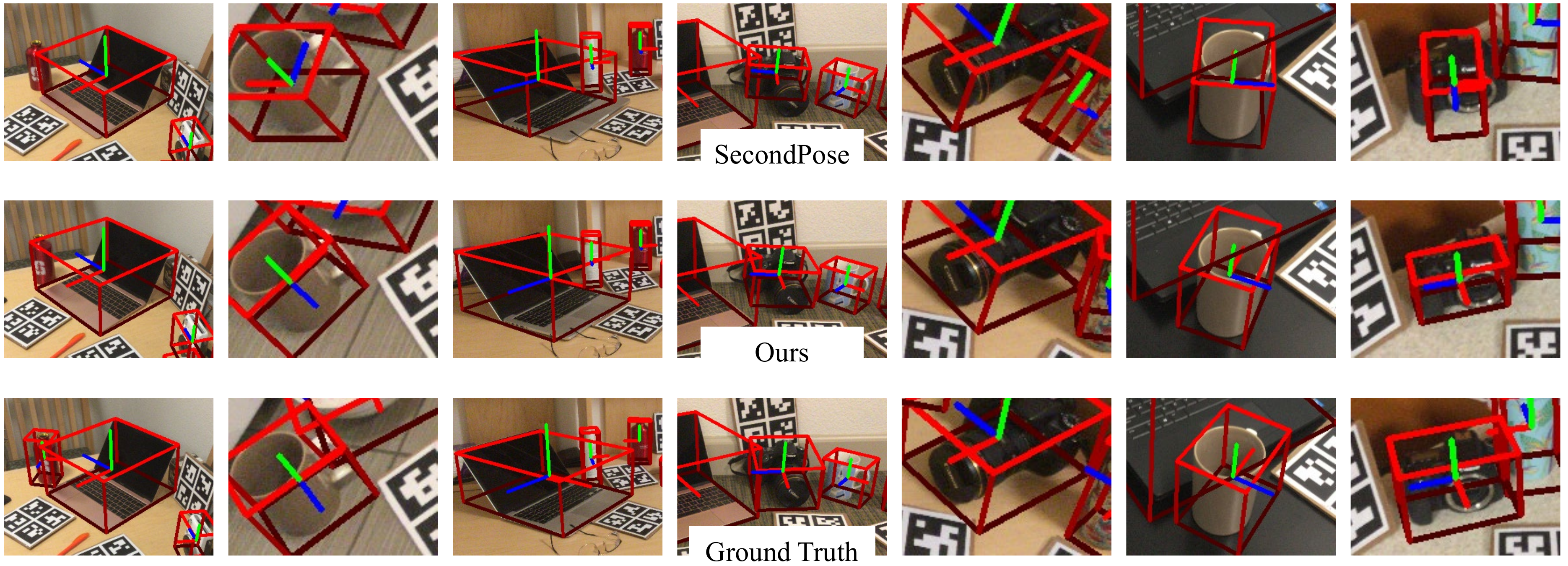}
	\caption{Qualitative results of 9D pose estimation on NOCS-REAL275. We  compare our predictions with ground-truth annotations and those produced by SecondPose. Our approach achieves markedly higher accuracy.} 
	\label{fig:real275}
\end{figure*}

\section{EXPERIMENTS}
We first evaluate our method on three widely used benchmarks against  SOTA to demonstrate its superiority. Next, visual experiments with real objects in various scenarios further demonstrate the effectiveness of our method. Finally, we integrate the proposed object pose estimation method into a designed picking system and conduct picking experiments on five object categories with four instances each, validating the effectiveness of our approach. Ablation studies further quantify each component's contribution.

\textit{Datasets}: We conduct experiments on the following  widely-used 9D pose estimation benchmarks: NOCS-REAL275~\cite{wang2019normalized}, NOCS-CAMERA25~\cite{wang2019normalized} and Wild6D\cite{fu2022category}. NOCS-REAL275 
consists of 13 real-world scenes featuring six object categories \textit{(bowl, bottle, can, camera, mug, and laptop)}, with 4,300 images from 7 scenes used for training and 2,750 images from the remaining 6 scenes for testing. NOCS-CAMERA25 is a synthetic dataset designed to enhance NOCS-REAL275. It includes 300,000 RGB-D images spanning 1,085 objects, with 25,000 images from 184 objects allocated for validation purposes. Wild6D comprises images from 486 videos, with only the test set annotated, covering 162 objects from five categories (excluding ``\textit{can}"). 

\textit{Metrics}: We employ the standard mean Average Precision (mAP (\%)) as the evaluation metric, computed for 
$5^{\circ}2cm$, $5^{\circ}5cm$, $10^{\circ}2cm$, and $10^{\circ}5cm$. Specifically, the metric denoted as $m^{\circ}ncm$ corresponds to the proportion of predictions where the rotation estimation error is constrained within $m$ degrees and the translation estimation error is limited to $n$ centimeters. Furthermore, we include the mAP value calculated at a 3D Intersection over Union (IoU) threshold of 75\% and 50\%, a metric that simultaneously assesses errors in both pose and size. The used metrics enable comprehensive evaluation. 

\textit{Implementation details}: In this paper, the raw images are provided at a resolution of \(640 \times 480\). We then utilize Grounded-SAM\cite{ren2024grounded} for segmentation in real-world and robotic picking evaluation. To ensure fairness in evaluating public datasets, we use Mask R-CNN \cite{he2017mask} for object segmentation, in line with other SOTA methods.  
During geometric feature extraction, we configure the hierarchical neighbor parameters as \(\mathcal{K}=[10, 20, 40, 80, 160, 300]\) and $C_{y}^{geo}$ equals 48 and 96 when \(y=1\) and \(y=2\). We calculate the appropriate bin for each point to be projected and utilize the sparsification feature in MinkowskiEngine\cite{choy20194d} to eliminate features that project into the same bin. Subsequently, we apply the densification feature in MinkowskiEngine\cite{choy20194d} to set unprojected bins to zero, which enables to automatically compute the gradients for this process. Our models are trained with batch size 48 on a single NVIDIA 4090 GPU up to epoch 40. The Adamw optimizer is employed with an initial learning rate set to 0.001. For learning rate scheduling, we use WarmupCosineLR with parameters including max iterations 200k, warmup factor 0.001, and warmup iterations 2K.

\begin{figure*}[!th]
	\centering
	\includegraphics[width=\textwidth]{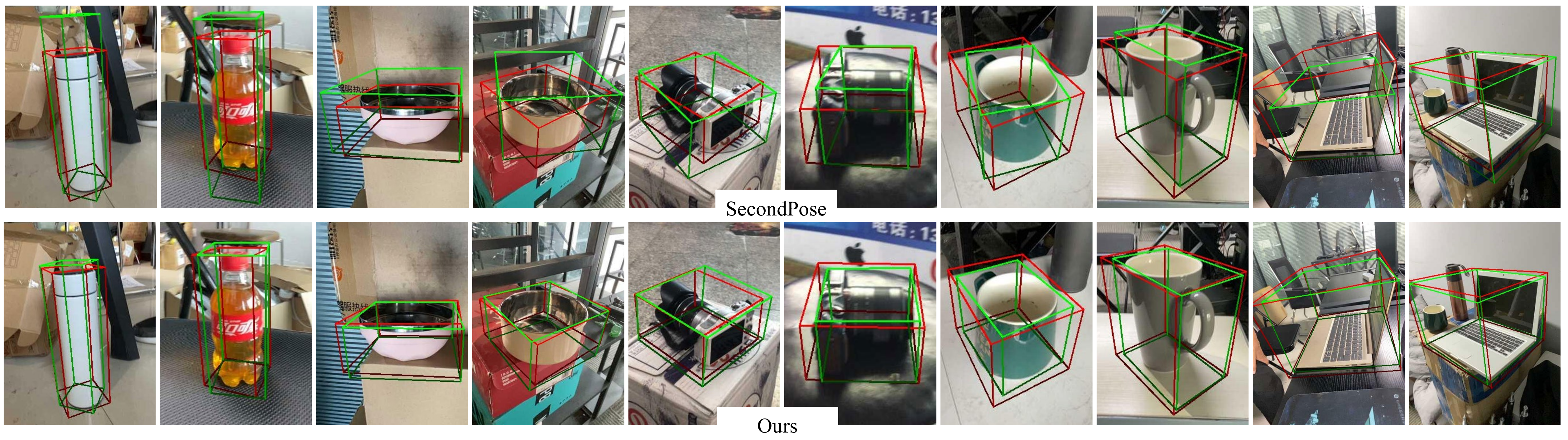}
	\caption{On the Wild6D dataset, we provide a comparative visualization of pose bounding boxes between our method and the baseline approach, SecondPose. Green and red correspond to the predicted results and ground truth, respectively. Specifically, the center of the 3D bounding box indicates the object's position, its orientation reflects the object's pose, and its length, width, and height represent the object's dimensions.}  
	\label{fig:wild6d}
\end{figure*}

\begin{table}[t]
\setlength{\tabcolsep}{5pt}
    \caption{Comparison of efficiency metrics on the NOCS-REAL275}
    \label{tab:efficiency_comparison}
    \centering
    \begin{tabular}{lcc}
        \toprule
        Method & Inference Speed (FPS) $\uparrow$ &  Parameter Size (MB) $\downarrow$ \\
        \midrule
        SecondPose\cite{chen2024secondpose} & 9.87 & 60.2 \\
        Ours & \textbf{10.16} & \textbf{48.2} \\
        \bottomrule
    \end{tabular}

\end{table}

\subsection{Evaluation on Public Datasets}

We conducted extensive experiments to validate our method on the NOCS-REAL275\cite{wang2019normalized}, NOCS-CAMERA25\cite{wang2019normalized}, and Wild6D\cite{fu2022category} datasets. Specifically, the experiments on NOCS-REAL275\cite{wang2019normalized} and Wild6D\cite{fu2022category} were trained on a mixed dataset where NOCS-REAL275\cite{wang2019normalized} and NOCS-CAMERA25 were combined at a ratio of 1:3, while the experiment on NOCS-CAMERA25\cite{wang2019normalized} was trained solely on the NOCS-CAMERA25\cite{wang2019normalized} dataset. The relevant quantitative results are presented in Table \ref{tab:1} and Table \ref{tab:wild6d}, where the best-performing results for each evaluation metric are highlighted in bold. The visualization results can be found in Fig. \ref{fig:real275} and Fig. \ref{fig:wild6d}.

\begin{table}[t]\centering
\caption{Quantitative results on the Wild6D dataset: comparison based on mean average precision (mAP) metrics of $IoU_{3D}$ (\%) and $m^\circ n\mathrm{cm}$ (\%)}
\label{tab:wild6d}
\setlength{\tabcolsep}{4pt}
\begin{tabular}{l|*{6}{>{\centering\arraybackslash}c}} 
\toprule
\multirow{2}{*}[-0.5\normalbaselineskip]{\makecell{Method}} & \multicolumn{6}{c}{Mean Average Precision (mAP)} \\ 
\cmidrule(lr){2-7}
& $IoU_{50}$ & $IoU_{75}$ & $5^\circ2\text{cm}$ & $5^\circ5\text{cm}$ & $10^\circ2\text{cm}$ & $10^\circ5\text{cm}$ \\ 
\midrule
NOCS\cite{wang2019normalized} & 0.00 & 0.00 & 0.04 & 0.04 & 0.04 & 0.04 \\
SPD\cite{tian2020shape} & 52.60 & 20.30 & 6.90 & 9.30 & 20.10 & 27.80 \\
SGPA\cite{chen2021sgpa} & 63.50 & 34.50 & 26.30 & 29.20 & 33.80 & 39.50 \\
GPV-Pose\cite{di2022gpv} & - & - & 14.10 & 21.50 & 23.80 & 41.10 \\
MH6D-sup\cite{liu2024mh6d} & 76.10 & \underline{41.90} & 27.00 & 31.20 & \underline{34.40} & 40.50 \\
Diff9D\cite{liu2025diff9d} & 76.48 & 38.16 & 25.52 & 30.46 & 32.48 & 40.94 \\
PENet\cite{yu2025category} & 74.60 & - & \underline{33.10} & 39.30 & - & 48.50 \\
SecondPose\cite{chen2024secondpose} & \textbf{77.48} & 36.04 & 27.56 & \underline{41.30} & 31.92 & \textbf{51.78} \\
Ours & \underline{77.34} & \textbf{43.04} & \textbf{33.30} & \textbf{42.32} & \textbf{38.56} & \underline{51.62} \\ 
\bottomrule
\end{tabular}
\end{table}

\subsubsection{NOCS-REAL275}
The left part of Table \ref{tab:1} presents a comprehensive quantitative comparison between our proposed method, the SOTA method of the same technical type (SecondPose\cite{chen2024secondpose} - highlighted in gray), and other SOTA approaches on the NOCS-REAL275\cite{wang2019normalized} dataset. In contrast to NOCS-CAMERA25\cite{wang2019normalized}, the data in NOCS-REAL275\cite{wang2019normalized} is entirely sourced from real-world scenarios, making the results on this dataset more indicative of the algorithm’s effectiveness in practical applications. Our method  outperforms all SOTA methods on NOCS-REAL275\cite{wang2019normalized}. Specifically, it exceeds SecondPose by 3.1 percentage points (pp) in the mAP at $5^\circ2\text{cm}$. Furthermore, our method remains competitive with GCE-Pose \cite{li2025gce}, a SOTA method that relies on shape priors, and even outperforms it by 2.3 pp in mAP over the entire test set under the most stringent $5^\circ2\text{cm}$ threshold. The qualitative comparison results are shown in Fig.~\ref{fig:real275}. Our method demonstrates higher accuracy in pose estimation, especially for the more challenging rotation information. Furthermore, in the case of the ``can'' at the right corner of the image in Column 5 of Fig. \ref{fig:real275}, our method still achieves success even when only a small amount of data is captured, whereas SecondPose\cite{chen2024secondpose} clearly makes an error in estimating the position of the can.

\begin{figure*}[!th]
	\centering
	\includegraphics[width=\textwidth]{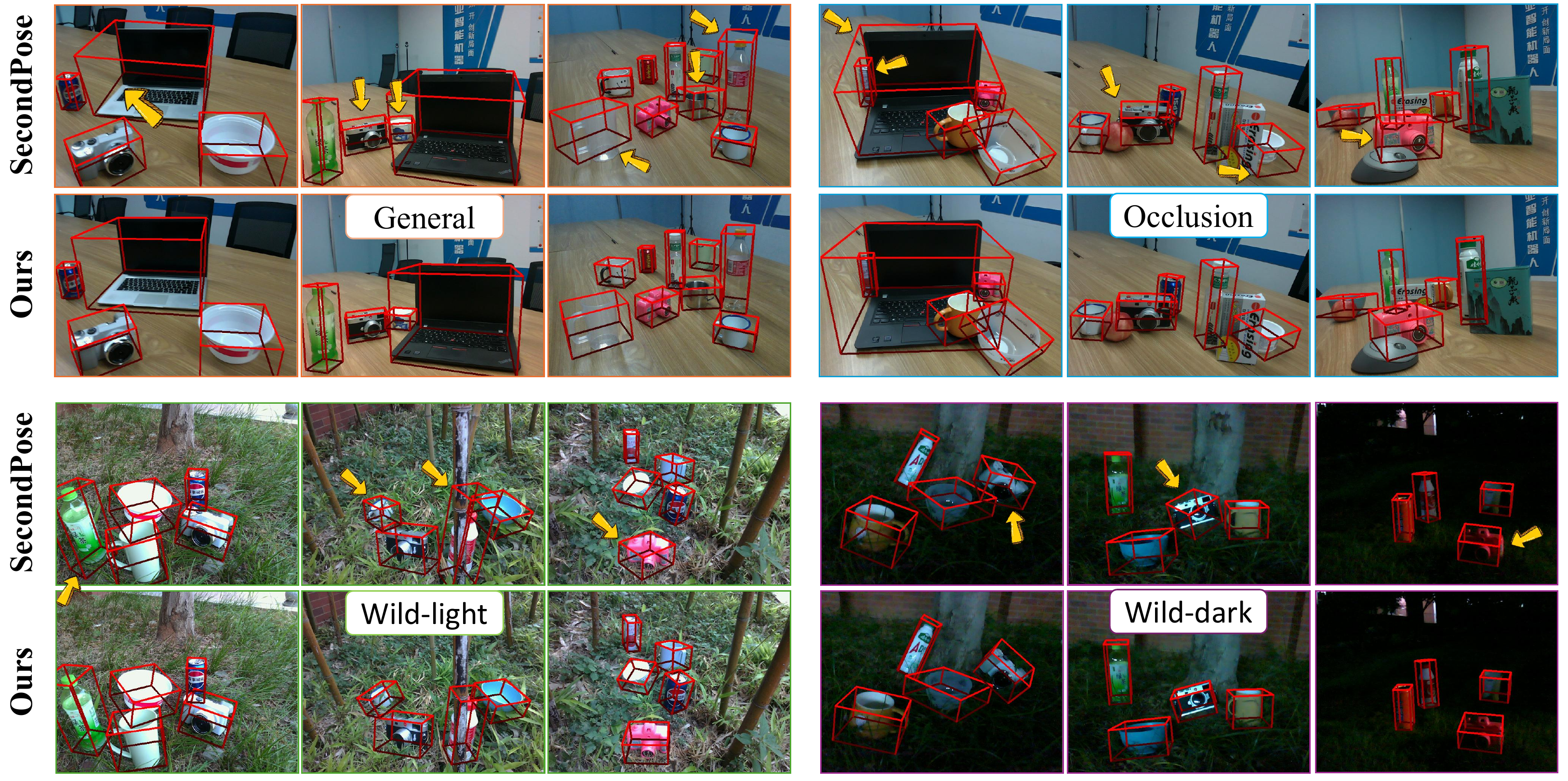}
	\caption{Qualitative results in real-world scenarios. We compare our method with SecondPose in four different real-world scenarios. To keep the result visualization clear and uncluttered, we hid the coordinate axes and only visualize the 3D bounding boxes. The length, width, and height of the 3D boxes reflect the size of the objects, their orientation indicates the object pose, and the center of the 3D boxes represents the object position. Arrows point to novel objects not present in the training set. Figure is best viewed  enlarged.}  
	\label{fig:shiwu}
\end{figure*}

\textit{Efficiency}: The efficiency of the proposed algorithm was evaluated using the NOCS-REAL275 test dataset, as shown in \ref{tab:efficiency_comparison}. To quantify the inference speed, a standard metric was adopted: the total time consumed during testing was divided by the total number of images in the test dataset, yielding the number of Frames processed Per Second (FPS). Note that each image here contains 5 to 6 objects.
Specifically, the baseline method (SecondPose) achieved an inference speed of 9.87 FPS with a model parameter size of 60.2 MB. In contrast, the proposed algorithm outperformed SecondPose in both key metrics: it attained a higher inference speed of 10.16 FPS while maintaining a more compact parameter size of 48.2 MB. Notably, this performance gain is substantial, as the proposed algorithm not only reduces the model complexity (via fewer parameters) but also simultaneously improves both inference accuracy and speed, which is non-trivial in the field of efficient computer vision model design.

\subsubsection{NOCS-CAMERA25}
The right part of Table \ref{tab:1} presents the quantitative experimental results of our method on the NOCS-CAMERA25\cite{wang2019normalized} dataset. The results demonstrate that our method significantly outperforms other methods in the same category across multiple metrics evaluating the accuracy of pose and size estimation on this dataset. Additionally, it still maintains a slight advantage in the mAP at $IoU_{75^*}$ and $5^\circ5\text{cm}$ compared to methods based on shape priors or correspondence. However, our method shows limited improvement and performs worse than correspondence-based methods at other thresholds. This can be attributed to the overly clean synthetic dataset, which allows the network to achieve good results despite lacking robustness. Additionally, the depth data is more accurate, enabling correspondence-based methods to better leverage their advantages. Notably, the experimental results on the synthetic data are only for reference and do not diminish the fact that our method still has advantages.

\begin{table}[t]
\setlength{\tabcolsep}{5pt}
    \caption{Success rate (\%) of robotic picking across five object classes}
    \label{tab:pick}
    \centering
    \begin{tabular}{c|ccccc|c}
        \toprule
        \multirow{2}{*}[-0.5\normalbaselineskip]{\makecell{Method}} & \multicolumn{6}{c}{Picking Success Rate (\%)} \\ 
\cmidrule(lr){2-7}
         & Bottle & Can & Bowl & Mug & Camera & Avg  \\
        \midrule
        SecondPose\cite{chen2024secondpose}   & 75   & 80 & 80 & 60 & 70   & 73   \\
        Ours       & \textbf{80}   & \textbf{90} & \textbf{95} & \textbf{75} & \textbf{90}   & \textbf{86}   \\
        \bottomrule
    \end{tabular}

\end{table}

\subsubsection{WILD6D}
Trained on the combined NOCS-CAMERA25\cite{wang2019normalized} and NOCS-REAL275\cite{wang2019normalized} datasets, our model was further evaluated on the Wild6D\cite{fu2022category} test set to validate its generalization capability, with results presented in Table \ref{tab:wild6d}. As evidenced in Table~\ref{tab:wild6d}, our method attains SOTA parity on the \(IoU_{75}\), \(5^{\circ}2\,\text{cm}\), \(5^{\circ}5\,\text{cm}\) and \(10^{\circ}5\,\text{cm}\) metrics, while also delivering strong performance on  the remaining metrics. Notably, it achieves  a 4.16 pp improvement on \(10^{\circ}2\,\text{cm}\). This indicates that our method still maintains its advantages on newly encountered data with unseen styles. The visualization results are provided in Fig.~\ref{fig:wild6d}, which displays comparative 9D pose visualizations of two distinct instances across five object categories: bottle, bowl, camera, mug, and laptop. Notably, for axially rotationally symmetric objects (e.g., bottles and bowls), rotational estimation is only meaningful with respect to the axis direction. It can be observed from the visualization results that our method exhibits greater advantages in terms of size and rotation estimation.

\subsection{Real-world and robotic picking evaluation}
To further verify the effectiveness and robustness of our method in estimating object poses in real-world scenarios, we conducted comparative visualization experiments between the baseline and our method across various real-world settings. Additionally, to validate the role of the method in practical applications, we designed a picking experiment.

{\hypersetup{urlcolor=blue}
\begin{figure*}[t]
	\centering
	\includegraphics[width=\textwidth]{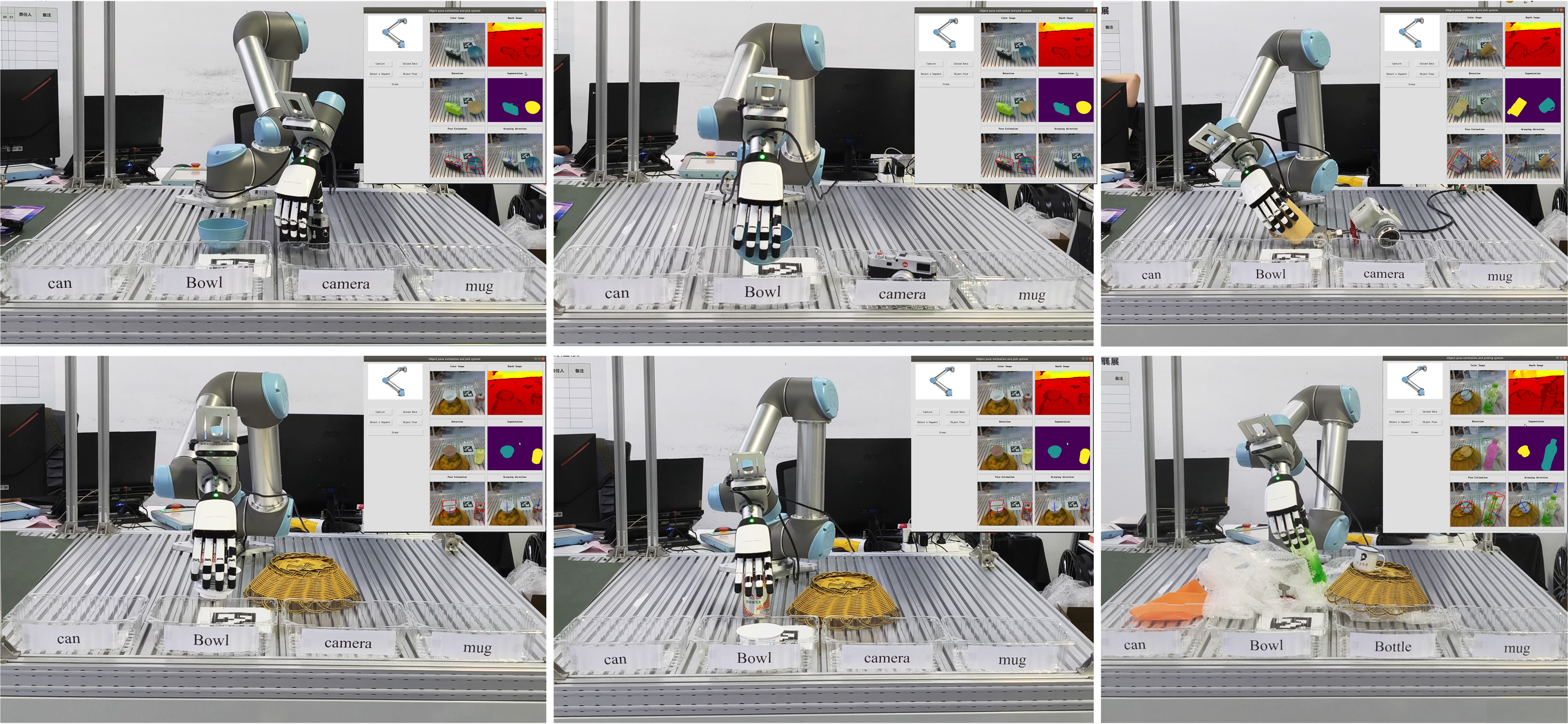}
	\caption{Category-level picking visualization, including five categories of objects: bottle, bowl, camera, can, and mug. The robotic picking demo can be seen at \url{https://youtu.be/z5LxH-4fI34}.}
	\label{fig:pick}

\end{figure*}
}

\subsubsection{Object Pose Estimation in the Real World}

We conducted extensive experiments across four scenarios: general, occlusion, wild-lighting, and dark environments. In this experiment, we  used the Intel RealSense L515 RGB-D camera as the data acquisition device and employed Grounded-SAM~\cite{ren2024grounded} for object segmentation. Notably, both the camera device and the segmentation algorithm can be replaced by any alternatives that achieve the same effect. As indicated by the yellow arrows in Fig. \ref{fig:shiwu}, our method remains effective for novel objects in real-world scenarios (none of the objects shown in Fig. \ref{fig:shiwu} were present in the training set). Furthermore, even in challenging scenarios such as occlusion, wild lighting, and dark environments, our method still outperforms the baseline. Additionally, it should be noted that in the third column of the ``general" section in Fig.~\ref{fig:shiwu}, the estimation of the mug with a smiling face failed. This is because the handle of that mug is pure black, and the camera used tends to produce abnormal depth data when capturing black regions.

\subsubsection{Robotic Picking Validation}

To verify the effectiveness of our method in practical industrial applications, we designed a robotic picking system. This system comprises a host computer, a UR5 robotic arm, a RH56E2-L dexterous hand, and an Intel RealSense D435i camera. Furthermore, as shown in Fig.~\ref{fig:pick}, we have designed a picking software system to serve as an interactive visualization interface. This system continues to employ Grounded-SAM~\cite{ren2024grounded} as the object segmentation module and our method as the pose estimation module, which is the core of the picking process. Picking poses are manually designed for each object category relative to the object's pose. To avoid mechanical interference, we enforce a technical setup such that only the back of the hand is visible in the presented images, , rather than the palm. We perform continuous picking in the order from near to far based on the distance of objects from the camera.

For five categories (bottle, can, camera, mug, bowl), we collected 4 different instances for each category and conducted 5 pick tests with different poses for each instance. Thus, a total of 100 picking experiments were performed for each method. It should be noted that all instances involved in the picking experiments have not been seen during the training. To focus on object pose, we consider the experiment successful as long as the object is picked up. As shown in Table \ref{tab:pick}, our method achieved an average picking success rate of 86\%, while the baseline reached 73\%, which demonstrates the superiority of our method.

\begin{table}[!t]
\setlength{\tabcolsep}{3pt}
    \caption{Ablation Studies of Network Configurations on REAL275}
    \label{tab:Abla}
    \centering
    \begin{tabular}{ccc|c|ccccc}
        \toprule
        \multicolumn{3}{c|}{\textbf{Module}} & \multirow{2}{*}{\textbf{$C_{geo}$}} & \multicolumn{5}{c}{\textbf{REAL275}} \\ \cmidrule(lr){1-3} \cmidrule(lr){5-9}
        \textbf{M\_1} & \textbf{M\_2} & \textbf{M\_3} &  & $IoU_{75^*}$ & $5^\circ2\text{cm}$ & $5^\circ5\text{cm}$ & $10^\circ2\text{cm}$ & $10^\circ5\text{cm}$ \\ \midrule
         & \checkmark & \checkmark & 96 & 50.9 & 57.2 & 65.5 & 75.4 & 86.6 \\
        \checkmark &  & \checkmark & 24 & 49.8 & 56.9 & 64.1 & 75.5 & 86.2 \\
        \checkmark & \checkmark &  & 96 & 51.4 & 58.6 & 65.8 & 76.7 & 86.9 \\ \midrule
        \checkmark & \checkmark & \checkmark & 24 & 51.4 & 58.2 & 64.5 & 76.4 & 86.4 \\
        \checkmark & \checkmark & \checkmark & 48 & 51.5 & 57.4 & 63.7 & 75.4 & 85.6 \\
        \checkmark & \checkmark & \checkmark & 96 & 51.7 & \textbf{59.3} & \textbf{66.3} & \textbf{77.2} & \textbf{87.4} \\
        \checkmark & \checkmark & \checkmark & 192 & 51.6 & 57.2 & 63.8 & 75.7 & 86.0 \\
        \checkmark & \checkmark & \checkmark & 384 & \textbf{51.8} & 55.8 & 62.2 & 75.6 & 85.8 \\ \bottomrule
    \end{tabular}

\end{table}

\subsection{Ablation Studies}

To further investigate the effectiveness of the constituent modules of the proposed method, we conducted extensive ablation experiments on the NOCS-Real275 dataset, which is more realistic and a widely used dataset.

\subsubsection{Module}
We conducted ablation studies for the constituents of our approach, and the  results are presented in Table~\ref{tab:Abla}. There are three key improvements in our method corresponding to three constituent modules. \textbf{M\_1}: The \textbf{SSA-Module} enhances the prediction accuracy of object translation and size by predicting the symmetric point of each point, and indirectly improves the rotation estimation accuracy. \textbf{M\_2}: The Sim(3)-consistent geometric features generated by the \textbf{MHP-Module} are more robust than the handcrafted geometric features in the baseline and more suitable for object pose estimation tasks.  \textbf{M\_3}: The \textbf{SlinConv block} improves the model's ability to model contextual information by implementing spherical convolution with a larger receptive field, thereby enhancing the algorithm accuracy. To study the role of each module, we replaced one of the three individually with the baseline’s corresponding module, with results in the upper part of Table \ref{tab:Abla}. No checkmark in the table indicates a replaced module. The results indicate that all three modules play a positive role in the network.

\subsubsection{Geometric feature dimension}
In computer vision tasks, feature dimensions are usually set to 192 or 384, which is more appropriate. Therefore, we further investigated the impact of geometric dimensions on the algorithm, and the results are shown in the lower part of Table \ref{tab:Abla}. We found that the algorithm performs best with a geometric feature dimension set to 96. Increasing the geometric feature dimension did not lead to significant performance improvements; rather, it resulted in a decline in performance and increased computational overhead. We attribute this to the fact that this feature is only an initial feature, and deeper features will be further learned and the dimension will be increased during spherical convolution. Moreover, the loss function does not explicitly supervise this feature, so a larger feature dimension here will make it difficult for the network to learn effective features.

\section{CONCLUSION AND FUTURE WORK}
This paper presents SSH-Pose, a shape prior-free framework for category-level 9D object pose estimation to enhance robotic picking. Key innovations include a symmetry-aware translation/size estimator using DINOv2 for symmetric geometry inference, and a novel rotation estimator that fuses DINOv2 features with our proposed efficient, parallelizable geometric features for rotation prediction. This rotation estimator is driven by our SlinConv block, which models contextual relationships without increasing computational cost to enhance rotation estimation. SSH-Pose achieves SOTA performance among shape prior-free and direct regression frameworks.
Extensive experiments on public datasets, real-world visualizations, and robotic picking tests validate its superiority. However, its training requires real-world data, which is scarce and hard to collect. Future work will focus on training solely with synthetic datasets and breaking domain barriers to reduce training requirements.

\bibliographystyle{IEEEtran}
\bibliography{ref}

@article{liu2024domain,
  title={Domain-generalized robotic picking via contrastive learning-based 6-d pose estimation},
  author={Liu, Jian and Sun, Wei and Yang, Hui and Liu, Chongpei and Zhang, Xing and Mian, Ajmal},
  journal={IEEE Transactions on Industrial Informatics},
  year={2024},
  publisher={IEEE}
}

@article{yu2025category,
  title={Category-Level 6-D Object Pose Estimation With Learnable Prior Embeddings for Robotic Grasping},
  author={Yu, Sheng and Zhai, Di-Hua and Yin, Jian and Xia, Yuanqing},
  journal={IEEE Transactions on Industrial Electronics},
  year={2025},
  publisher={IEEE}
}

@inproceedings{lin2022sar,
  title={Sar-net: Shape alignment and recovery network for category-level 6d object pose and size estimation},
  author={Lin, Haitao and Liu, Zichang and Cheang, Chilam and Fu, Yanwei and Guo, Guodong and Xue, Xiangyang},
  booktitle={Proceedings of the IEEE/CVF conference on computer vision and pattern recognition},
  pages={6707--6717},
  year={2022}
}

@inproceedings{liu2023net,
  title={Ist-net: Prior-free category-level pose estimation with implicit space transformation},
  author={Liu, Jianhui and Chen, Yukang and Ye, Xiaoqing and Qi, Xiaojuan},
  booktitle={Proceedings of the IEEE/CVF International Conference on Computer Vision},
  pages={13978--13988},
  year={2023}
}

@inproceedings{lin2023vi,
  title={Vi-net: Boosting category-level 6d object pose estimation via learning decoupled rotations on the spherical representations},
  author={Lin, Jiehong and Wei, Zewei and Zhang, Yabin and Jia, Kui},
  booktitle={Proceedings of the IEEE/CVF international conference on computer vision},
  pages={14001--14011},
  year={2023}
}

@inproceedings{chen2024secondpose,
  title={Secondpose: Se (3)-consistent dual-stream feature fusion for category-level pose estimation},
  author={Chen, Yamei and Di, Yan and Zhai, Guangyao and Manhardt, Fabian and Zhang, Chenyangguang and Zhang, Ruida and Tombari, Federico and Navab, Nassir and Busam, Benjamin},
  booktitle={Proceedings of the IEEE/CVF Conference on Computer Vision and Pattern Recognition},
  pages={9959--9969},
  year={2024}
}

@article{oquab2023dinov2,
  title={Dinov2: Learning robust visual features without supervision},
  author={Oquab, Maxime and Darcet, Timoth{\'e}e and Moutakanni, Th{\'e}o and Vo, Huy and Szafraniec, Marc and Khalidov, Vasil and Fernandez, Pierre and Haziza, Daniel and Massa, Francisco and El-Nouby, Alaaeldin and others},
  journal={arXiv preprint arXiv:2304.07193},
  year={2023}
}

@article{qi2017pointnet++,
  title={Pointnet++: Deep hierarchical feature learning on point sets in a metric space},
  author={Qi, Charles Ruizhongtai and Yi, Li and Su, Hao and Guibas, Leonidas J},
  journal={Advances in neural information processing systems},
  volume={30},
  year={2017}
}

@inproceedings{tian2020shape,
  title={Shape prior deformation for categorical 6d object pose and size estimation},
  author={Tian, Meng and Ang Jr, Marcelo H and Lee, Gim Hee},
  booktitle={European Conference on Computer Vision},
  pages={530--546},
  year={2020},
  organization={Springer}
}

@inproceedings{wang2019normalized,
  title={Normalized object coordinate space for category-level 6d object pose and size estimation},
  author={Wang, He and Sridhar, Srinath and Huang, Jingwei and Valentin, Julien and Song, Shuran and Guibas, Leonidas J},
  booktitle={Proceedings of the IEEE/CVF conference on computer vision and pattern recognition},
  pages={2642--2651},
  year={2019}
}

@article{umeyama2002least,
  title={Least-squares estimation of transformation parameters between two point patterns},
  author={Umeyama, Shinji},
  journal={IEEE Transactions on pattern analysis and machine intelligence},
  volume={13},
  number={4},
  pages={376--380},
  year={2002},
  publisher={IEEE}
}

@inproceedings{lin2022category,
  title={Category-level 6d object pose and size estimation using self-supervised deep prior deformation networks},
  author={Lin, Jiehong and Wei, Zewei and Ding, Changxing and Jia, Kui},
  booktitle={European Conference on Computer Vision},
  pages={19--34},
  year={2022},
  organization={Springer}
}

@article{liu2024mh6d,
  title={Mh6d: Multi-hypothesis consistency learning for category-level 6-d object pose estimation},
  author={Liu, Jian and Sun, Wei and Liu, Chongpei and Yang, Hui and Zhang, Xing and Mian, Ajmal},
  journal={IEEE Transactions on Neural Networks and Learning Systems},
  year={2024},
  publisher={IEEE}
}

@inproceedings{lin2021dualposenet,
  title={Dualposenet: Category-level 6d object pose and size estimation using dual pose network with refined learning of pose consistency},
  author={Lin, Jiehong and Wei, Zewei and Li, Zhihao and Xu, Songcen and Jia, Kui and Li, Yuanqing},
  booktitle={Proceedings of the IEEE/CVF international conference on computer vision},
  pages={3560--3569},
  year={2021}
}

@inproceedings{di2022gpv,
  title={Gpv-pose: Category-level object pose estimation via geometry-guided point-wise voting},
  author={Di, Yan and Zhang, Ruida and Lou, Zhiqiang and Manhardt, Fabian and Ji, Xiangyang and Navab, Nassir and Tombari, Federico},
  booktitle={Proceedings of the IEEE/CVF Conference on Computer Vision and Pattern Recognition},
  pages={6781--6791},
  year={2022}
}

@inproceedings{zheng2023hs,
  title={Hs-pose: Hybrid scope feature extraction for category-level object pose estimation},
  author={Zheng, Linfang and Wang, Chen and Sun, Yinghan and Dasgupta, Esha and Chen, Hua and Leonardis, Ale{\v{s}} and Zhang, Wei and Chang, Hyung Jin},
  booktitle={Proceedings of the IEEE/CVF conference on computer vision and pattern recognition},
  pages={17163--17173},
  year={2023}
}

@inproceedings{he2017mask,
  title={Mask r-cnn},
  author={He, Kaiming and Gkioxari, Georgia and Doll{\'a}r, Piotr and Girshick, Ross},
  booktitle={Proceedings of the IEEE international conference on computer vision},
  pages={2961--2969},
  year={2017}
}

@inproceedings{choy20194d,
  title={4d spatio-temporal convnets: Minkowski convolutional neural networks},
  author={Choy, Christopher and Gwak, JunYoung and Savarese, Silvio},
  booktitle={Proceedings of the IEEE/CVF conference on computer vision and pattern recognition},
  pages={3075--3084},
  year={2019}
}

@inproceedings{chen2021sgpa,
  title={Sgpa: Structure-guided prior adaptation for category-level 6d object pose estimation},
  author={Chen, Kai and Dou, Qi},
  booktitle={Proceedings of the IEEE/CVF international conference on computer vision},
  pages={2773--2782},
  year={2021}
}

@inproceedings{li2025gce,
  title={Gce-pose: Global context enhancement for category-level object pose estimation},
  author={Li, Weihang and Xu, Hongli and Huang, Junwen and Jung, Hyunjun and Yu, Peter KT and Navab, Nassir and Busam, Benjamin},
  booktitle={Proceedings of the Computer Vision and Pattern Recognition Conference},
  pages={27154--27165},
  year={2025}
}

@inproceedings{irshad2022centersnap,
  title={Centersnap: Single-shot multi-object 3d shape reconstruction and categorical 6d pose and size estimation},
  author={Irshad, Muhammad Zubair and Kollar, Thomas and Laskey, Michael and Stone, Kevin and Kira, Zsolt},
  booktitle={2022 International Conference on Robotics and Automation (ICRA)},
  pages={10632--10640},
  year={2022},
  organization={IEEE}
}

@inproceedings{lin2024instance,
  title={Instance-adaptive and geometric-aware keypoint learning for category-level 6d object pose estimation},
  author={Lin, Xiao and Yang, Wenfei and Gao, Yuan and Zhang, Tianzhu},
  booktitle={Proceedings of the IEEE/CVF Conference on Computer Vision and Pattern Recognition},
  pages={21040--21049},
  year={2024}
}

@article{fu2022category,
  title={Category-level 6d object pose estimation in the wild: A semi-supervised learning approach and a new dataset},
  author={Fu, Yang and Wang, Xiaolong},
  journal={Advances in Neural Information Processing Systems},
  volume={35},
  pages={27469--27483},
  year={2022}
}

@article{ren2024grounded,
  title={Grounded sam: Assembling open-world models for diverse visual tasks},
  author={Ren, Tianhe and Liu, Shilong and Zeng, Ailing and Lin, Jing and Li, Kunchang and Cao, He and Chen, Jiayu and Huang, Xinyu and Chen, Yukang and Yan, Feng and others},
  journal={arXiv preprint arXiv:2401.14159},
  year={2024}
}

@article{liu2025diff9d,
  title={Diff9d: Diffusion-based domain-generalized category-level 9-dof object pose estimation},
  author={Liu, Jian and Sun, Wei and Yang, Hui and Deng, Pengchao and Liu, Chongpei and Sebe, Nicu and Rahmani, Hossein and Mian, Ajmal},
  journal={IEEE Transactions on Pattern Analysis and Machine Intelligence},
  year={2025},
  publisher={IEEE}
}

@book{dekking2005modern,
  title={A Modern Introduction to Probability and Statistics: Understanding why and how},
  author={Dekking, Frederik Michel},
  year={2005},
  publisher={Springer Science \& Business Media}
}

@inproceedings{liu2022catre,
  title={Catre: Iterative point clouds alignment for category-level object pose refinement},
  author={Liu, Xingyu and Wang, Gu and Li, Yi and Ji, Xiangyang},
  booktitle={European Conference on Computer Vision},
  pages={499--516},
  year={2022},
  organization={Springer}
}

@inproceedings{yu2024inceptionnext,
  title={Inceptionnext: When inception meets convnext},
  author={Yu, Weihao and Zhou, Pan and Yan, Shuicheng and Wang, Xinchao},
  booktitle={Proceedings of the IEEE/cvf conference on computer vision and pattern recognition},
  pages={5672--5683},
  year={2024}
}

@article{ren2025learning,
  title={Learning shape-independent transformation via spherical representations for category-level object pose estimation},
  author={Ren, Huan and Yang, Wenfei and Liu, Xiang and Zhang, Shifeng and Zhang, Tianzhu},
  journal={arXiv preprint arXiv:2503.13926},
  year={2025}
}

@article{zhou2025multiscale,
  title={Multiscale-Hashing Network for 6-D Pose Estimation of Unseen Objects in the Wild},
  author={Zhou, Jiaming and Zhu, Qing and Wang, Yaonan and Feng, Mingtao and Liu, Xuebing and Shafait, Faisal and Mian, Ajmal},
  journal={IEEE/ASME Transactions on Mechatronics},
  year={2025},
  publisher={IEEE}
}

@article{yu2023robotic,
  title={Robotic grasp detection based on category-level object pose estimation with self-supervised learning},
  author={Yu, Sheng and Zhai, Di-Hua and Xia, Yuanqing},
  journal={IEEE/ASME Transactions on Mechatronics},
  volume={29},
  number={1},
  pages={625--635},
  year={2023},
  publisher={IEEE}
}

@article{liang2024adaptive,
  title={Adaptive human--robot interaction torque estimation with high accuracy and strong tracking ability for a lower limb rehabilitation robot},
  author={Liang, Xu and Yan, Yuchen and Wang, Weiqun and Su, Tingting and He, Guangping and Li, Guotao and Hou, Zeng-Guang},
  journal={IEEE/ASME Transactions on Mechatronics},
  volume={29},
  number={6},
  pages={4814--4825},
  year={2024},
  publisher={IEEE}
}

@article{he2023contourpose,
  title={ContourPose: Monocular 6-D pose estimation method for reflective textureless metal parts},
  author={He, Zaixing and Li, Quanzhi and Zhao, Xinyue and Wang, Jin and Shen, Huarong and Zhang, Shuyou and Tan, Jianrong},
  journal={IEEE Transactions on Robotics},
  volume={39},
  number={5},
  pages={4037--4050},
  year={2023},
  publisher={IEEE}
}

@article{chen2020edge,
  title={Edge-dependent efficient grasp rectangle search in robotic grasp detection},
  author={Chen, Lu and Huang, Panfeng and Li, Yuanhao and Meng, Zhongjie},
  journal={IEEE/ASME Transactions on Mechatronics},
  volume={26},
  number={6},
  pages={2922--2931},
  year={2020},
  publisher={IEEE}
}

@inproceedings{liu2023gsnet,
  title={Gsnet: Model reconstruction network for category-level 6d object pose and size estimation},
  author={Liu, Penglei and Zhang, Qieshi and Cheng, Jun},
  booktitle={2023 IEEE International Conference on Robotics and Automation (ICRA)},
  pages={2898--2904},
  year={2023},
  organization={IEEE}
}

@inproceedings{lin2020convolution,
  title={Convolution in the cloud: Learning deformable kernels in 3d graph convolution networks for point cloud analysis},
  author={Lin, Zhi-Hao and Huang, Sheng-Yu and Wang, Yu-Chiang Frank},
  booktitle={Proceedings of the IEEE/CVF conference on computer vision and pattern recognition},
  pages={1800--1809},
  year={2020}
}

@article{zhang2025learning,
  title={Learning Cross-View Consistent 3D Keypoints for Object 6D Pose Estimation},
  author={Zhang, Shaobo and Zhao, Wanqing and Guan, Ziyu and Zhao, Wei and Peng, Jinye and Fan, Jianping},
  journal={IEEE Transactions on Circuits and Systems for Video Technology},
  year={2025},
  publisher={IEEE}
}

@article{zhou2025canonical,
  title={Canonical Shape Reconstruction with SE (3) Equivariance Learning for Weakly-Supervised Object Pose Estimation},
  author={Zhou, Jun and Chen, Kai and Wei, Mingqiang and Zhang, Xiao-Ping and Dou, Qi and Qin, Jing},
  journal={IEEE Transactions on Circuits and Systems for Video Technology},
  year={2025},
  publisher={IEEE}
}

@article{liu2022hff6d,
  title={HFF6D: Hierarchical feature fusion network for robust 6D object pose tracking},
  author={Liu, Jian and Sun, Wei and Liu, Chongpei and Zhang, Xing and Fan, Shimeng and Wu, Wei},
  journal={IEEE Transactions on Circuits and Systems for Video Technology},
  volume={32},
  number={11},
  pages={7719--7731},
  year={2022},
  publisher={IEEE}
}

@inproceedings{su2019deep,
  title={Deep multi-state object pose estimation for augmented reality assembly},
  author={Su, Yongzhi and Rambach, Jason and Minaskan, Nareg and Lesur, Paul and Pagani, Alain and Stricker, Didier},
  booktitle={2019 IEEE International Symposium on Mixed and Augmented Reality Adjunct (ISMAR-Adjunct)},
  pages={222--227},
  year={2019},
  organization={IEEE}
}

@article{marchand2015pose,
  title={Pose estimation for augmented reality: a hands-on survey},
  author={Marchand, Eric and Uchiyama, Hideaki and Spindler, Fabien},
  journal={IEEE transactions on visualization and computer graphics},
  volume={22},
  number={12},
  pages={2633--2651},
  year={2015},
  publisher={IEEE}
}

@article{lin2024clipose,
  title={Clipose: Category-level object pose estimation with pre-trained vision-language knowledge},
  author={Lin, Xiao and Zhu, Minghao and Dang, Ronghao and Zhou, Guangliang and Shu, Shaolong and Lin, Feng and Liu, Chengju and Chen, Qijun},
  journal={IEEE Transactions on Circuits and Systems for Video Technology},
  volume={34},
  number={10},
  pages={9125--9138},
  year={2024},
  publisher={IEEE}
}

@article{zou2023gpt,
  title={Gpt-cope: A graph-guided point transformer for category-level object pose estimation},
  author={Zou, Lu and Huang, Zhangjin and Gu, Naijie and Wang, Guoping},
  journal={IEEE Transactions on Circuits and Systems for Video Technology},
  volume={34},
  number={4},
  pages={2385--2398},
  year={2023},
  publisher={IEEE}
}

@article{tang20193d,
  title={3D mapping and 6D pose computation for real time augmented reality on cylindrical objects},
  author={Tang, Fulin and Wu, Yihong and Hou, Xiaohui and Ling, Haibin},
  journal={IEEE Transactions on Circuits and Systems for Video Technology},
  volume={30},
  number={9},
  pages={2887--2899},
  year={2019},
  publisher={IEEE}
}

@article{kim2002object,
  title={Object-based video abstraction for video surveillance systems},
  author={Kim, Changick and Hwang, Jenq-Neng},
  journal={IEEE Transactions on Circuits and Systems for Video Technology},
  volume={12},
  number={12},
  pages={1128--1138},
  year={2002},
  publisher={IEEE}
}

@ARTICLE{SinRef6D,
  author={Liu, Jian and Sun, Wei and Zeng, Kai and Zheng, Jin and Yang, Hui and Rahmani, Hossein and Mian, Ajmal and Wang, Lin},
  journal={IEEE Transactions on Robotics}, 
  title={Scalable Unseen Objects 6-DoF Absolute Pose Estimation With Robotic Integration}, 
  year={2026},
  volume={42},
  number={},
  pages={1884-1901}
}

@article{liu2026survey,
  title={Deep Learning-Based Object Pose Estimation: A Comprehensive Survey},
  author={Liu, Jian and Sun, Wei and Yang, Hui and Zeng, Zhiwen and Liu, Chongpei and Zheng, Jin and Liu, Xingyu and Rahmani, Hossein and Sebe, Nicu and Mian, Ajmal},  
  journal={International Journal of Computer Vision},
  year={2026},
  volume={134},
  number={81},
  pages={1-45}
}

\begin{IEEEbiography}[{\includegraphics[width=1in,height=1.25in,clip,keepaspectratio]{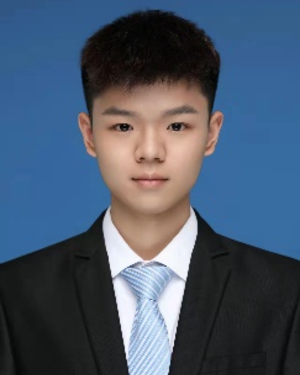}}]{Panfei Cheng}
received the B.S. degree from Hunan University, Changsha, China, in 2022. He is currently
working toward a Ph.D. degree with the National
Engineering Laboratory for Robot Visual Perception
and Control, College of Electrical And Information
Engineering, Hunan University, China, under the supervision of Prof. Hongshan Yu. His research interests include object pose estimation, point cloud registration and robotic manipulation.
\end{IEEEbiography}

\begin{IEEEbiography}[{\includegraphics[width=1in,height=1.25in,clip,keepaspectratio]{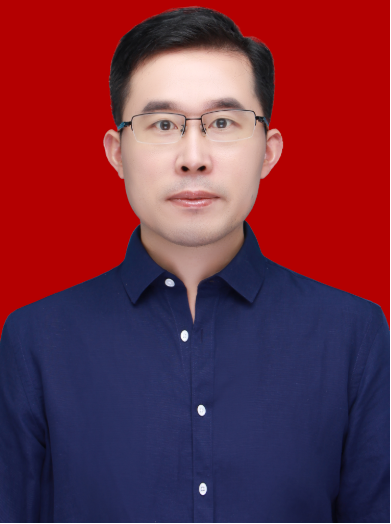}}]{Hongshan Yu}
received the B.S., M.S., and Ph.D.
degrees in control science and technology in electrical and information engineering from Hunan University, Changsha, China in 2001, 2004, and 2007,respectively. From 2011 to 2012, he was a Post-Doctoral Researcher with the Laboratory for Computational Neuroscience, University of Pittsburgh,
USA. He is currently a Professor and Associate Dean of the School of Robotics and Artificial Intelligence at Hunan University.

Prof. Yu was the Associate Dean of the National Engineering Laboratory for Robot Visual Perception and Control during 2016-2020 and received two Second-Grade National Science and Technology Progress Awards of China in 2004 and 2006. His research interests include computer vision, robotics, and artificial intelligence, with over 50 related publications.
\end{IEEEbiography}

\begin{IEEEbiography}[{\includegraphics[width=1in,height=1.25in,clip,keepaspectratio]{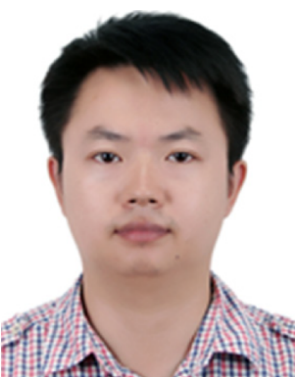}}]{Wenrui Chen}
received the B.S.
and Ph.D. degrees in mechanical engineering from
the Huazhong University of Science and Technology,
Wuhan, China, in 2010 and 2017, respectively.

In 2016, he was a Visiting Scholar with the
University of Lincoln, Lincoln, U.K. Since 2017, he
has been with the School of Artificial Intelligence
and Robotics, Hunan University, Changsha, China,
where he is currently an Associate Professor. His
research interests include robotic multimodal perception,
dexterous manipulation, and skill learning.

\end{IEEEbiography}
\begin{IEEEbiography}[{\includegraphics[width=1in,height=1.25in,clip,keepaspectratio]{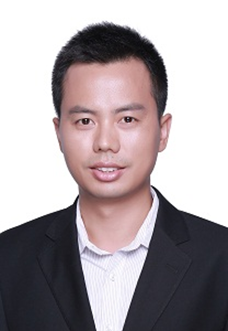}}]{Xiaojun Tang}
received the B.S. and Ph.D. degrees from Harbin Engineering University in 2010 and 2015, respectively. Now he is a researcher at China Academy of Space Technology.

Dr. Tang has published over 70 refereed papers in journals and conferences. His research interests include the aerospace engineering, space environment test, visual inspection and micro deformation measurement in complex environment.

\end{IEEEbiography}

\begin{IEEEbiography}[{\includegraphics[width=1in,height=1.25in,clip,keepaspectratio]{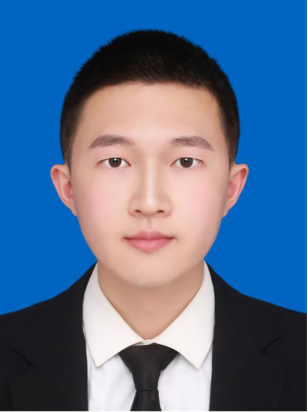}}]{Jian Liu}
received his Ph.D. degree
at the National Engineering Research Center of
Robot Visual Perception and Control Technology, College of Electrical and Information Engineering, Hunan University, Changsha, China, in 2025. From 2023 to 2024, he was a Visiting Scholar at the Department of Computer Science of the University of Western Australia, Perth, WA, Australia, under the supervision of Prof. Ajmal Mian. His current research interests include 3D computer vision, object pose estimation, and robotic manipulation. He served as a reviewer for more than 20 journals and conferences, including the IEEE TPAMI, IEEE TIP, IEEE TNNLS, IEEE TMECH, IEEE TII, and IEEE ICRA/IROS, etc.

\end{IEEEbiography}

\begin{IEEEbiography}[{\includegraphics[width=1in,height=1.25in,clip,keepaspectratio]{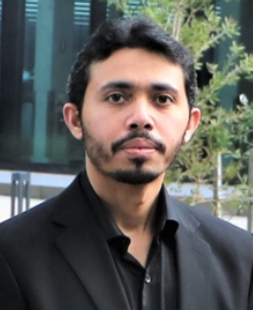}}]{Naveed Akhtar}
is currently an ARC DECRA Fellow at the School of Computing and Information Systems, The University of Melbourne. He has previously served as a Sr. Lecturer at the University of Western Australia and as a Research Fellow at the Australian National University. He has served as an Area Chair of prestigious conferences in computer vision, including IEEE CVPR, ECCV, IEEE WACV. He currently serves as an Associate Editor for IEEE Trans. on Neural Networks and Learning Systems. His research has contributed towards interpretability of deep learning visual models, adversarial attacks and defenses for deep learning, point cloud analysis, medical image analysis and hyper/multi-spectral image analysis. The outcomes have led to acknowledgements and honors such as Google Research Program Award, ACM Distinguished Speaker and Western Australia’s Early Career Scientist of the Year 2021 Finalist.

\end{IEEEbiography}
\vfill

\end{document}